\documentclass[conference]{IEEEtran}
\IEEEoverridecommandlockouts
\usepackage{cite}
\usepackage{textcomp}
\def\BibTeX{{\rm B\kern-.05em{\sc i\kern-.025em b}\kern-.08em
    T\kern-.1667em\lower.7ex\hbox{E}\kern-.125emX}}

\usepackage{times}

\usepackage[dvipsnames]{xcolor}
\usepackage[normalem]{ulem}
\usepackage{lineno}
\usepackage[bookmarks=false]{hyperref}

\usepackage{graphicx}
\usepackage{subfigure} 
\usepackage{bbold} 
\usepackage{amsmath,amssymb,mathrsfs,amsthm} 
\usepackage{mathtools} 

\usepackage{algorithm}
\usepackage{algpseudocode}
\usepackage{multirow}
\usepackage{gensymb}
\usepackage{enumitem} 
\usepackage{siunitx}

\usepackage{array}

\usepackage{nomencl}
\makenomenclature

\newcolumntype{L}{>{\centering\arraybackslash}m{3cm}}

\DeclareMathOperator{\Tr}{Tr}

\begin{document}

\title{Estimate Three-Phase Distribution Line Parameters With Physics-Informed Graphical Learning Method}

\author{Wenyu~Wang,~\IEEEmembership{Student~Member,~IEEE,} Nanpeng~Yu,~\IEEEmembership{Senior~Member,~IEEE,}
\thanks{W. Wang and N. Yu are with the Department of Electrical and Computer Engineering, University of California, Riverside, CA 92521, USA. Email: nyu@ece.ucr.edu.}}


\maketitle

\begin{abstract}
Accurate estimates of network parameters are essential for modeling, monitoring, and control in power distribution systems. In this paper, we develop a physics-informed graphical learning algorithm to estimate network parameters of three-phase power distribution systems. Our proposed algorithm uses only readily available smart meter data to estimate the three-phase series resistance and reactance of the primary distribution line segments. We first develop a parametric physics-based model to replace the black-box deep neural networks in the conventional graphical neural network (GNN). Then we derive the gradient of the loss function with respect to the network parameters and use stochastic gradient descent (SGD) to estimate the physical parameters. Prior knowledge of network parameters is also considered to further improve the accuracy of estimation. Comprehensive numerical study results show that our proposed algorithm yields high accuracy and outperforms existing methods.
\end{abstract}


\nomenclature[1E]{$E^{(i,j)}_{m \times n}$}{An $m\times n$ matrix, in which the $ij$-th element is 1 and the rest of elements are all 0.}
\nomenclature[1Im]{$Im(\cdot)$}{Imaginary part of a complex variable.}
\nomenclature[1Re]{$Re(\cdot)$}{Real part of a complex variable.}
\nomenclature[1M]{$M$}{Number of loads in a circuit.}
\nomenclature[1N]{$N$, $\mathfrak{L}$}{Number of non-substation nodes and lines in the three-phase primary distribution network.}
\nomenclature[1T]{$\mathfrak{T}$}{A batch of time difference instances.}
\nomenclature[1c]{$\textrm{co}(n)$, $\textrm{ne}(n)$, $\textrm{no}(m)$}{Edge set connected to node $n$, node neighbor set of node $n$, node that meter $m$ connects to.}
\nomenclature[1p]{$p_n^i$, $q_n^i$, $\alpha_n^i$, $\beta_n^i$}{Real and imaginary part of power injection, real and imaginary part of node $n$'s voltage at phase $i$.}
\nomenclature[1u]{$\boldsymbol{u}$, $\boldsymbol{s}$}{$3\times 1$ Vector of a node's three-phase complex voltage and complex power injection.}
\nomenclature[1v]{$v$}{Voltage magnitude scalar of a smart meter.}
\nomenclature[1v1]{$\tilde{v}$, $\tilde{o}$}{Time difference of $v$ and $o$.}
\nomenclature[1w]{$\boldsymbol{w}$}{The set of parameters of the GNN/distribution system.}
\nomenclature[1x]{$\boldsymbol{x}$, $\boldsymbol{l}$, $o$}{State vector, feature vector, and output scalar of an element in the GNN.}
\nomenclature[1x1]{$[\boldsymbol{x}]$, $[\boldsymbol{l}]$, $[o]$}{A vector that stacks all the states, features, and output.}
\nomenclature[1x2]{$\hat{\boldsymbol{x}}_n(t)$, $\hat{\boldsymbol{l}}_n(t)$}{$12\times 1$ vectors that stack $\boldsymbol{x}_n(t-1)$ and $\boldsymbol{x}_n(t)$, $\boldsymbol{l}_n(t-1)$ and $\boldsymbol{l}_n(t)$, respectively.}
\nomenclature[1x3]{$[\hat{\boldsymbol{x}}(t)]$, $[\hat{\boldsymbol{l}}(t)]$}{Vectors of stacking $[\boldsymbol{x}(t-1)]$ and $[\boldsymbol{x}(t)]$, $[\boldsymbol{l}(t-1)]$ and $[\boldsymbol{l}(t)]$, respectively.}
\nomenclature[3i]{$(\cdot)^i$}{A variable in phase $i$.}
\nomenclature[3i]{$(\cdot)^{ij}$}{A variable between phase $i$ and $j$.}
\nomenclature[3n]{$(\cdot)_n$}{A variable of the $n$-th node, load, or element.}
\nomenclature[3t]{$\cdot (t)$}{The value of a variable at time $t$.}
\nomenclature[4o]{$\odot$, $\oslash$}{Element-wise multiplication and division.}
\nomenclature[4o]{$\mathbb{0}_{m\times n}$}{All-0 matrix of size $m\times n$.}
\printnomenclature

\section{Introduction}
Accurate modeling of three-phase power distribution systems is crucial to accommodating the increasing penetration of distributed energy resources (DERs). To monitor and coordinate the operations of DERs, several key applications such as three-phase power flow, state estimation, optimal power flow, and network reconfiguration are needed. All of these depend on accurate three-phase distribution network models, which include the network topology and parameters \cite{wang2016phase}. However, the distribution network topology and parameters in the geographic information system (GIS) often contain errors because the model documentation usually becomes unreliable during the system modifications and upgrades \cite{foggo2019improving}.

Although topology estimation for distribution networks has been studied extensively \cite{liao2018unbalanced,wang2020maximum}, the estimation of distribution network parameters such as line impedances still needs further development. It is more challenging to estimate parameters of power distribution networks than that of transmission networks. This is because the distribution lines are rarely transposed, which lead to unequal diagonal and off-diagonal elements in the impedance matrix. Thus, three-phase line models need to be developed instead of single-phase equivalent models. Specifically, the elements of the $3 \times 3$ phase impedance matrix need to be estimated for each three-phase line segment.

Many methods have been proposed to estimate transmission network parameters. However, very few of them can be applied to the three-phase distribution networks using readily available sensor data. The existing parameter estimation literature can be roughly classified into three groups based on the type of sensor data used. 

In the first group of literature, supervisory control and data acquisition (SCADA) system data such as power and current injections are used to estimate transmission network parameters of a single-phase model. Most of the algorithms in this group perform joint state and parameter estimation by residual sensitivity analysis and state vector augmentation \cite{zarco2000power}. Parameter errors are detected using identification indices \cite{logic2005approach,castillo2010offline}, enhanced normalized Lagrange multipliers \cite{lin2016enhancing}, and projection statistics \cite{zhao2018robust}. Adaptive data selection \cite{li2017measurement} is used to improve parameter estimation accuracy.



In the second group of literature, phasor measurement unit (PMU) data such as voltage and current phasors are used to estimate line parameters of transmission and distribution systems \cite{asprou2015estimation,kumar2016state,khandeparkar2016detection,gajare2017method,ren2017tracking,yu2018patopaem,ardakanian2019identification}. Although these methods achieve highly accurate parameter estimates, they require costly and widespread installation of PMUs. Linear least squares is used to estimate transmission line parameters \cite{asprou2015estimation}. Parallel Kalman filter for a bilinear model is used to estimate both states and line parameters of the transmission system \cite{kumar2016state}. With single-phase transmission line models, nonlinear least squares is used to estimate line parameters and calibrate remote meters \cite{khandeparkar2016detection}. Traveling waves are used to estimate parameters of series compensated lines \cite{gajare2017method}. An augmented state estimation method is developed to estimate three-phase transmission line parameters \cite{ren2017tracking}. Maximum likelihood estimation (MLE) is used to estimate single-phase distribution line parameters \cite{yu2018patopaem}. Lasso is adopted to estimate three-phase admittance matrix in distribution systems \cite{ardakanian2019identification}.



In the third group of literature, smart meter data such as voltage magnitude and complex power consumption are used to estimate distribution line parameters \cite{han2015automated,peppanen2016distribution,zhang2020topology,cunha2020automated,wang2020parameter}. Particle swarm \cite{han2015automated} and linear regression  \cite{lave2019distribution,zhang2020topology} are used to estimate single-phase line parameters. Linear approximation of voltage drop \cite{peppanen2016distribution} is used to estimate the parameters of single-phase and balanced three-phase distribution lines. Multiple linear regression model is used to estimate three-phase line impedance in \cite{cunha2020automated}, but it does not work with delta-connected smart meters with phase-to-phase measurement. In \cite{wang2020parameter}, three-phase line parameters are estimated through MLE based on a linearized physical model.

The existing methods for parameter estimation either assume a single-phase equivalent distribution network model or require widespread installation of micro-PMUs, which are cost prohibitive. To fill the knowledge gap, this paper develops a physics-informed graphical learning algorithm to estimate the $3 \! \times \! 3$ series resistance and reactance matrices of three-phase distribution line model using readily available smart meter measurements. Our proposed method is inspired by the emerging graph neural network (GNN), which is designed for estimation problems in networked systems. We develop three-phase power flow-based physical transition functions to replace the ones based on deep neural networks in the GNN. We then derive the gradient of the voltage magnitude loss function with respect to the line segments' resistance and reactance parameters with an iterative method. Finally, the estimates of distribution network parameters can be updated with the stochastic gradient descent (SGD) approach to minimize the error between the physics-based graph learning model and the smart meter measurements. Prior estimates and bounds of network parameters are also leveraged to improve the estimation accuracy. To improve computation efficiency, partitions can be introduced so that parameter estimations are executed in parallel in sub-networks.

The main technical contributions of this work are:
\begin{itemize}
  \item A physics-informed graphical learning method is developed to estimate line parameters of three-phase distribution networks.
  \item Our proposed algorithm only uses readily available smart meter data and can be easily applied to real-world distribution circuits.
  \item By preserving the nonlinearity of three-phase power flows in the graphical learning framework, our proposed approach yields more accurate parameter estimates on test feeders than the state-of-the-art benchmark.
\end{itemize}

The rest of the paper is organized as follows. Section \ref{sec:setup_assumption} describes the problem setup and assumptions. Section \ref{sec:frame_GNNintro} presents the overall framework of the proposed method and briefly introduces the GNN. Section \ref{sec:method_detail} provides the technical methods for construction and parameter estimation based on the physics-informed graphical model. Section \ref{sec:numerical_study} evaluates the performance of the proposed algorithm with a comprehensive numerical study. Section \ref{sec:conclusion} states the conclusion.

\section{Problem Setup and Assumptions} \label{sec:setup_assumption}

\subsection{Problem Setup}
The objective of this work is to estimate the series resistance and reactance in the $3\!\times\!3$ phase impedance matrix of three-phase primary lines of a distribution feeder. The impedance matrix of a line $l$ can be written as, $Z_l\!=\!R_l\!+\!j\!X_l$, where
\begin{equation}\label{eq:problem_setup_1}
R_l
\triangleq
\begin{bmatrix}
r_l^{aa} & r_l^{ab} & r_l^{ac} \\
r_l^{ab} & r_l^{bb} & r_l^{bc} \\
r_l^{ac} & r_l^{bc} & r_l^{cc} 
\end{bmatrix}, \quad
X_l
\triangleq
j \begin{bmatrix}
x_l^{aa} & x_l^{ab} & x_l^{ac} \\
x_l^{ab} & x_l^{bb} & x_l^{bc} \\
x_l^{ac} & x_l^{bc} & x_l^{cc} 
\end{bmatrix}.
\end{equation}
Since $Z_l$ is symmetric, for each line segment there are 6 resistance and 6 reactance parameters. The network contains $\mathfrak{L}$ lines and $N+1$ nodes, indexed as node $0$ to $N$. Node $0$ is the source node (e.g., a substation). In total, there are $12\mathfrak{L}$ parameters to estimate. $M$ loads are connected to the primary lines through the non-source nodes. The loads can be single-phase, two-phase, or three-phase.

\subsection{Assumptions} \label{sec:assumption}
The assumptions of measurement data and the network model are summarized below. First, for a single-phase load on phase $i$, the smart meter records real and reactive power injections and voltage magnitude of phase $i$. Second, for a two-phase delta-connected load between phase $i$ and $j$, the smart meter records the power injection and voltage magnitude across phase $i$ and $j$. Third, for a three-phase load, the smart meter records total power injection and voltage magnitude of a known phase $i$. Fourth, SCADA system records the voltage measurements at the source node. Fifth, it is assumed that the phase connections of all loads are known. Sixth, the topology of the primary three-phase feeder is known. Seventh, we assume that the GIS contains rough estimates of the network parameters. Assumptions one to four are based on the typical measurement configurations of smart meters and SCADA. Assumptions five to seven are based on the available information in GIS. 

\section{Overall Framework and Review of the GNN} \label{sec:frame_GNNintro}
\subsection{Overall Framework of the Proposed Method}
\begin{figure}[tbh]
    \centering
    \includegraphics[width=0.48\textwidth]{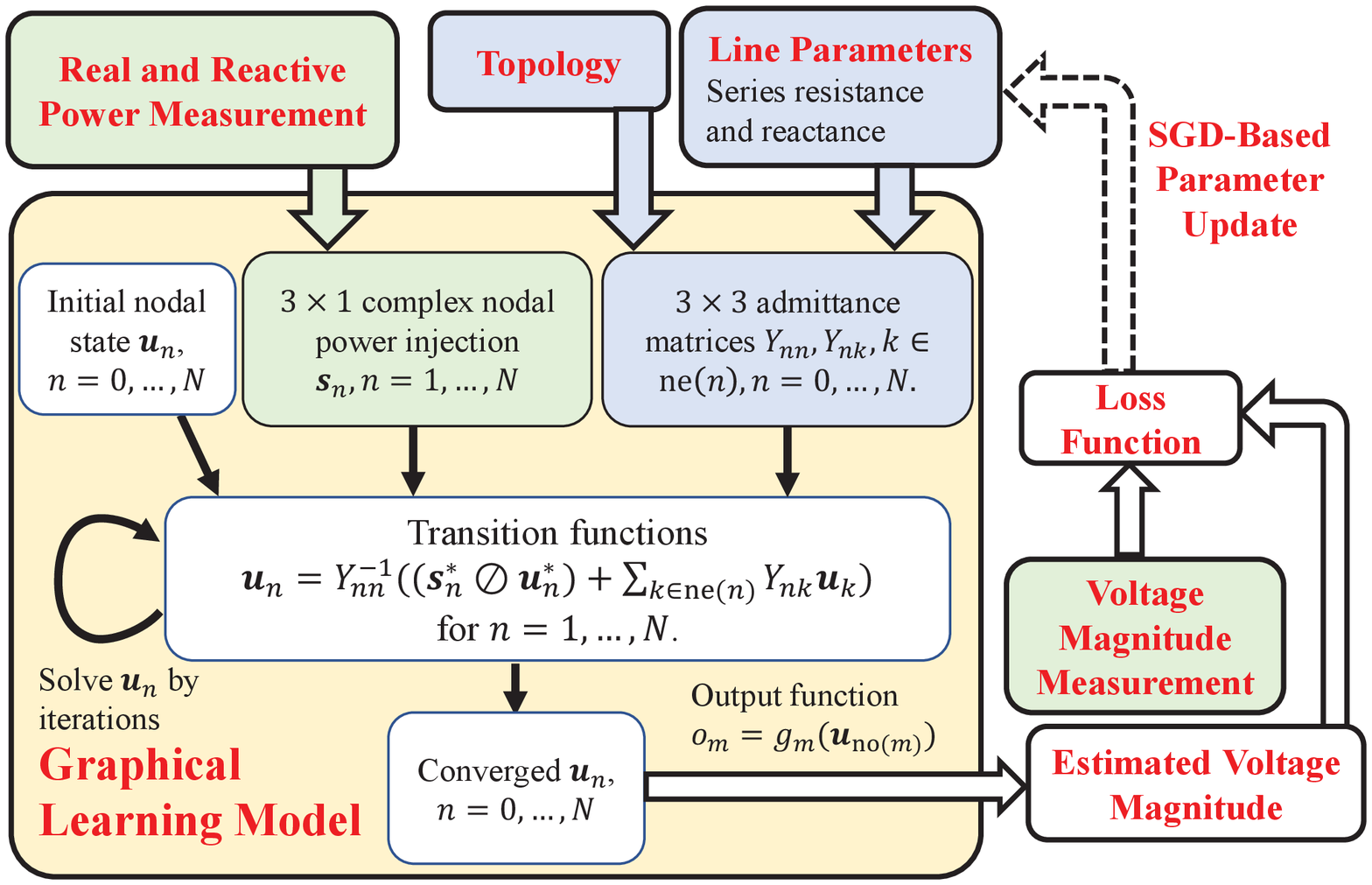}
    \caption{Framework of the method. The bold boxes with red titles represent higher-level elements. Green boxes represent smart meter data, and blue boxes represent distribution network information.}
    \label{fig:method_framework}
\end{figure}

The overall framework of the proposed graphical learning method for distribution line parameter estimation is illustrated in Fig. \ref{fig:method_framework}. As shown in the figure, a physics-informed graphical learning engine is constructed based on nonlinear power flow. The inputs to the graphical learning engine include power injection measurements from smart meters, distribution network topology, and distribution line parameters. In the graphical learning engine, each node corresponds to a physical bus in the distribution network. The nodal states, i.e., the three-phase complex voltage are iteratively updated by a set of transition functions. The graphical learning engine's outputs are the estimated smart meter voltage magnitudes, which are used to calculate the graphical learning engine's loss function. The gradient of the line parameters is computed from the loss function and subsequently used to update the line parameters using stochastic gradient descent. The technical details of the proposed method will be explained in \ref{sec:method_detail}.

\subsection{A Brief Overview of the GNN}
A GNN is a neural network model, which uses a graph's topological relationships between nodes to incorporate the underlying graph-structured information in data \cite{scarselli2008graph}. GNNs have been successfully applied in many different domains, such as social networks, image processing, and chemistry \cite{zhou2018graph}. Our proposed physics-informed graphical learning model is developed by embedding physics of power distribution networks into the standard GNN.

\begin{figure}[tbh]
    \centering
    \includegraphics[width=0.35\textwidth]{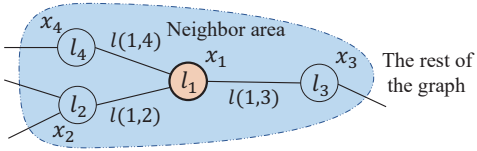}
    \caption{An illustration of a node and its neighbor area in a GNN.}
    \label{fig:GNN_example}
\end{figure}

The GNN is comprised of nodes connected by edges. The nodes represent objects or concepts, and the edges represent the relationships between nodes. Two vectors are attached to a node $n$: the state vector $\boldsymbol{x}_n$ and the feature vector $\boldsymbol{l}_n$. A feature vector $\boldsymbol{l}_{(m,n)}$ is attached to edge $(m,n)$. The state $\boldsymbol{x}_n$, which embeds information from its neighborhood with arbitrary depth, is naturally defined by the features of itself and the neighboring nodes and edges through a local parametric transition function $f_{\boldsymbol{w},n}$. A local output $o_n$ of node $n$, representing a local decision, is produced through a parametric output function $g_{\boldsymbol{w},n}$. The local transition and output functions are defined as follows:
\begin{equation}\label{eq:GNN_intro-1}
\begin{aligned}
\boldsymbol{x}_n &= f_{\boldsymbol{w},n}(\boldsymbol{l}_n,\boldsymbol{l}_{\textrm{co}(n)},\boldsymbol{x}_{\textrm{ne}(n)},\boldsymbol{l}_{\textrm{ne}(n)})\\
o_n &= g_{\boldsymbol{w},n}(\boldsymbol{x}_n,\boldsymbol{l}_n)
\end{aligned}
\end{equation}
Here, $\boldsymbol{l}_{\textrm{co}(n)}$, $\boldsymbol{x}_{\textrm{ne}(n)}$, and $\boldsymbol{l}_{\textrm{ne}(n)}$ are the features of edges connected to node $n$, the states of node $n$'s neighbor nodes, and the features of node $n$'s neighbor nodes. $\boldsymbol{w}$ is the set of parameters defining the transition and output functions. An example of a node and its neighbor area in a GNN is depicted in Fig. \ref{fig:GNN_example}. The local transition function for node $1$ is $\boldsymbol{x}_1 =f_{\boldsymbol{w},1}(\boldsymbol{l}_1,\boldsymbol{l}_{(1,2)},\boldsymbol{l}_{(1,3)},\boldsymbol{l}_{(1,4)},\boldsymbol{x}_2,\boldsymbol{x}_3,\boldsymbol{x}_4,\boldsymbol{l}_2,\boldsymbol{l}_3,\boldsymbol{l}_4)$. The implementation of the transition and output functions are flexible. They can be modeled as linear or nonlinear functions (e.g., neural networks).
Let $[\boldsymbol{x}]$, $[o]$, $[\boldsymbol{l}]$, and $[\boldsymbol{l}_N]$ represent the vectors constructed by stacking all the states, all the outputs, all the features, and all the node features, respectively. Then \eqref{eq:GNN_intro-1} can be represented in a compact form:
\begin{equation}\label{eq:GNN_intro-2}
\begin{aligned}
[\boldsymbol{x}] &=F_{\boldsymbol{w}}([\boldsymbol{x}],[\boldsymbol{l}])\\
[o] &=G_{\boldsymbol{w}}([\boldsymbol{x}],[\boldsymbol{l}_N])
\end{aligned}
\end{equation}
Here, $F_{\boldsymbol{w}}$ and $G_{\boldsymbol{w}}$ are the global transition function and global output function, which stacks all nodes' $f_{\boldsymbol{w},n}$ and $g_{\boldsymbol{w},n}$, respectively.

With the sufficient condition provided by the Banach fixed point theorem \cite{khamsi2011introduction}, one can find a unique solution of the state $[\boldsymbol{x}]$ for \eqref{eq:GNN_intro-2} using the classic iterative scheme:
\begin{equation}\label{eq:GNN_intro-3}
[\boldsymbol{x}]^{\tau+1}=F_{\boldsymbol{w}}([\boldsymbol{x}]^{\tau},[\boldsymbol{l}])
\end{equation}
Here, $[\boldsymbol{x}]^{\tau}$ is the $\tau$-th iteration of $[\boldsymbol{x}]$. The dynamic system of \eqref{eq:GNN_intro-3} converges exponentially fast to the solution of system \eqref{eq:GNN_intro-2} for any initial value $[\boldsymbol{x}]^0$.

The parameters $\boldsymbol{w}$ of a GNN's global transition and output functions $F_{\boldsymbol{w}}$ and $G_{\boldsymbol{w}}$ are updated and learned such that the output $[o]$ approximate the target values, i.e., minimizing a quadratic loss function:
\begin{equation}\label{eq:GNN_intro-4}
loss= \sum_{m=1}^M (o_m-\check{o}_m)^2
\end{equation}
Here, $M$ is the number of elements (number of measurements) in $[o]$, and  $o_m$ and $\check{o}_m$ are the $m$-th output and target value. The learning algorithm is based on a gradient-descent strategy. Since the iterative scheme in \eqref{eq:GNN_intro-3} is equivalent to a recurrent neural network, the gradient is calculated in a more efficient approach based on the Almeida-Pineda algorithm. Additional technical details of the GNN can be found in \cite{scarselli2008graph,pineda1987generalization,almeida1990learning}.

\section{Technical Methods} \label{sec:method_detail}
This section is organized as follows. Section \ref{sec:construct_transition} describes the construction of transition function $F_{\boldsymbol{w}}$. Section \ref{sec:construct_output} describes the formulation of the output function $G_{\boldsymbol{w}}$ and the loss function. Section \ref{sec:construct_grad} derives the gradient of the loss function. The use of prior knowledge of line parameters is described in Section \ref{sec:use_prior}. Section \ref{sec:learning_algorithm} presents the parameter estimation algorithm. The network partition method, which improves the scalability of the algorithm is described in Section \ref{sec:partition}. 

Our proposed physics-informed graphical learning model is different from the GNN \cite{scarselli2008graph}. In the GNN, $F_{\boldsymbol{w}}$ and $G_{\boldsymbol{w}}$ are often represented by neural networks whose weights are being learned. However, in our proposed framework, $F_{\boldsymbol{w}}$ and $G_{\boldsymbol{w}}$ are built based on the physical model of the power distribution systems. The parameters to be estimated are the line resistance and reactance. 
\subsection{Construction of the Transition Function}\label{sec:construct_transition}
The transition function is constructed based on the nonlinear power flow model of the distribution system. Let $\boldsymbol{s}_n \triangleq [s_n^a,s_n^b,s_n^c]^T$ be a $3\!\times\!1$ vector of nodal three-phase complex power injection of node $n$. $s_n^i \triangleq p_n^i+jq_n^i, i=a,b,c$, where $p_n^i$ and $q_n^i$ are node $n$'s real and reactive power injection of phase $i$. $\boldsymbol{s}_n$ can be derived from smart meters' power consumption data and phase connections as described in Section III-A of \cite{wang2020maximum}. Similarly, we define three-phase complex nodal voltage as $\boldsymbol{u}_n \triangleq [u_n^a,u_n^b,u_n^c]^T$, $u_n^i \triangleq \alpha_n^i+j\beta_n^i, i=a,b,c$. Let $Y_{nk}\!=\!Z_{nk}^{-1}$ be the $3\!\times\!3$ admittance matrix of the line between node $n$ and $k$, which can be calculated by using the topology and line parameters of the distribution network. Ignoring the negligible shunt, the three-phase power flow equation of node $n$ can be written as:
\begin{equation}\label{eq:transition-1}
\boldsymbol{s}_n=\boldsymbol{u}_n \odot \Big( Y_{nn}^* \boldsymbol{u}_n^*- \sum_{k\in \textrm{ne}(n)} Y_{nk}^* \boldsymbol{u}_k^*\Big)
\end{equation}
Here, $Y_{nn}=\sum_{k\in \textrm{ne}(n)} Y_{nk}$, $\odot$ is the element-wise multiplication, $\textrm{ne}(n)$ is the set of $n$'s neighbor nodes, and $(\cdot)^*$ represents complex conjugate. An equivalent form of \eqref{eq:transition-1} is:
\begin{equation}\label{eq:transition-2}
\boldsymbol{u}_n=Y_{nn}^{-1}\Big((\boldsymbol{s}_n^*\oslash \boldsymbol{u}_n^*)+\sum_{k\in ne(n)} Y_{nk} \boldsymbol{u}_k \Big)
\end{equation}
Here, $\oslash$ represents element-wise division. 

Next we convert \eqref{eq:transition-2} from a complex equation to a real-valued equation. For a matrix $A$, we define
\begin{equation}\label{eq:transition-3}
\langle A \rangle \triangleq \begin{bmatrix}\!
Re(A) & -Im(A)  \\
Im(A) & Re(A)
\!\end{bmatrix}
\end{equation}
Here, $Re(A)$ and $Im(A)$ are the real and imaginary part of $A$. Then, \eqref{eq:transition-2} can be rewritten as the local transition function:
\begin{equation}\label{eq:transition-4}
\begin{bmatrix} \!
Re\!(\! \boldsymbol{u}_n \! )  \\
Im\!(\! \boldsymbol{u}_n \! ) 
\! \end{bmatrix} \! =\!\langle \! Z_{nn} \! \rangle \!
\bigg(\!
\begin{bmatrix} \!
Re(\! \boldsymbol{s}_n^* \! \oslash \! \boldsymbol{u}_n^* \! )  \\
Im(\! \boldsymbol{s}_n^* \! \oslash \! \boldsymbol{u}_n^* \! ) 
\! \end{bmatrix}\! +\!\sum_{k\! \in \! \textrm{ne}\!(\!n\!)} \! \langle \! Y_{nk} \! \rangle \!
\begin{bmatrix} \!
Re\!(\!\boldsymbol{u}_k \! )  \\
Im\!(\! \boldsymbol{u}_k \!)
\! \end{bmatrix}
\! \bigg)
\end{equation}
Here $Z_{nn}\!\triangleq \! Y_{nn}^{-1}$. We define $6 \! \times \! 1$ state vector $\boldsymbol{x}_n$ and feature vector $\boldsymbol{l}_n$ of node $n$ as
\begin{equation}\label{eq:transition-5}
\boldsymbol{x}_n \triangleq
\begin{bmatrix}
Re(\boldsymbol{u}_n)  \\
Im(\boldsymbol{u}_n) 
\end{bmatrix} , \ 
\boldsymbol{l}_n \triangleq
\begin{bmatrix}
Re(\boldsymbol{s}_n)  \\
Im(\boldsymbol{s}_n) 
\end{bmatrix}
\end{equation}

Now, we can convert the local transition function \eqref{eq:transition-4} into the standard form and the global compact form:
\begin{equation}\label{eq:transition-6}
\begin{aligned}
\boldsymbol{x}_n & =f_{\boldsymbol{w},n}(\boldsymbol{x}_n,\boldsymbol{l}_n,\boldsymbol{x}_{\textrm{ne}(n)}) \ \textrm{(local form of node $n$)}\\
[\boldsymbol{x}] &=F_{\boldsymbol{w}}([\boldsymbol{x}],[\boldsymbol{l}]) \ \textrm{(global compact form)}
\end{aligned}
\end{equation}
For each node in a distribution system, we can derive a local transition function and stack them to obtain the global form of $F_{\boldsymbol{w}}$ as in \eqref{eq:transition-6}. Note that $[\boldsymbol{l}]$ only contains all the nodes' features and does not contain any edge features. The model's parameter $\boldsymbol{w}$ is the set of all lines' three-phase resistance and reactance, which is embedded in $\langle Z_{nn} \rangle$ and $\langle Y_{nk} \rangle$ of \eqref{eq:transition-4}.


Given line parameter $\boldsymbol{w}$, we can calculate the theoretical node state values of each time instance $t$ by iteratively applying the transition function \eqref{eq:transition-6}. This iteration procedure is formulated as a function called FORWARD shown in Algorithm \ref{algorithm:forward}. In the algorithm, step 1 initializes all nodes' states. In step 2, the global transition function is constructed. Step 3--6 estimate the nodes' states iteratively, while $\boldsymbol{x}_0(t)$ is fixed to its initial value because it is the measurement at the reference node. The iteration continues until convergence, which is controlled by a small ratio $\epsilon_\textrm{{forward}}$.
\begin{algorithm}[tbh]
\caption{FORWARD($\boldsymbol{w}$, $t$)} \label{algorithm:forward}
\begin{algorithmic}[1]
\Require Current line parameter $\boldsymbol{w}$ and the time instance $t$.
\Ensure Theoretical $[\boldsymbol{x}(t)]$ of the distribution system with line parameter $\boldsymbol{w}$.
\State Initialize the source nodes' state $\boldsymbol{x}_0(t)$ with the known measurement at the source node. Initialize the other nodes' state $\boldsymbol{x}_n(t)$ as defined in \eqref{eq:transition-5} with balanced flat node voltage, i.e. $\boldsymbol{u}_n(t)=[1,e^{-j\frac{2\pi}{3}},e^{j\frac{2\pi}{3}}]^T$, $(n=1,...,N)$.
\State Construct the initial $[\boldsymbol{x}(t)]^0$ by stacking all the initial $\boldsymbol{x}_n(t)$, $(n=0,...,N)$. Construct function $F_{\boldsymbol{w}}$ with $\boldsymbol{w}$.
\Repeat
\State $[\boldsymbol{x}(t)]^{\tau+1}=F_{\boldsymbol{w}}([\boldsymbol{x}(t)]^{\tau},[\boldsymbol{l}(t)])$ and fix $\boldsymbol{x}_0(t)$ to its initial value.
\State $\tau=\tau+1$
\Until{$\Vert [\boldsymbol{x}(t)]^{\tau}-[\boldsymbol{x}(t)]^{\tau-1} \Vert^2 < \epsilon_\textrm{{forward}} \cdot \Vert [\boldsymbol{x}(t)]^{\tau-1} \Vert^2 $}
\State \Return $[\boldsymbol{x}(t)]=[\boldsymbol{x}(t)]^{\tau}$.
\end{algorithmic}
\end{algorithm}

\subsection{Construction of the Output and Loss Function} \label{sec:construct_output}
The output of our proposed graphical learning model is the estimated smart meters' voltage measurements. For smart meter $m$, the estimated output $o_m$ is in the form of:
\begin{equation} \label{eq:output_1}
\begin{aligned}
o_m & =g_m(\boldsymbol{x}_{\textrm{no}(m)}) \ \textrm{(local form of meter $m$)}\\
[o] &=G([\boldsymbol{x}]) \ \textrm{(global compact form)}
\end{aligned}
\end{equation}
Here, $\boldsymbol{x}_{\textrm{no}(m)}$ is the state of the node, which the smart meter $m$ is connected to. Suppose we have a solution of the state $[\boldsymbol{x}(t)]=\textrm{FORWARD}(\boldsymbol{w},t)$, then $[o(t)]=G([\boldsymbol{x}(t)])$. Though $\boldsymbol{x}_{\textrm{no}(m)}$ has 6 elements from 3 phases, a smart meter only measures one single-phase or one phase-phase voltage magnitude. Based on the assumptions in Section \ref{sec:assumption}, if $k=\textrm{no}(m)$, then $g_m$ is defined as follows:
\begin{equation}\label{eq:output_2}
g_m \! (\! \boldsymbol{x}_k \!) \! = \!
\begin{cases}
\begin{aligned}
 & \sqrt{\! (\alpha_k^i)^2 \! + \! (\beta_k^i)^2} \quad \textrm{if meter $m$ is single-phase or}\\
 & \qquad\qquad\qquad \textrm{three-phase, measuring phase $i$} \\
 & \sqrt{\! (\alpha_k^i \! - \!\alpha_k^j)^2 \! + \! (\beta_k^i \! - \! \beta_k^j)^2} \quad \textrm{if meter $m$ is}\\
 & \qquad\qquad\qquad \textrm{two-phase, measuring phase $ij$}
\end{aligned}
\end{cases}
\end{equation}
Note that in the line parameter estimation formulation, $g_m$ does not depend on the parameter vector $\boldsymbol{w}$. Thus, it is not a parametric function.

Next we derive the loss function. To remove trends, instead of directly using the voltage output $[o]$, we use the first difference of the output time series. The estimated first difference of output time series for meter $m$ is:
\begin{equation}\label{eq:loss_2}
\tilde{o}_m(t)\triangleq o_m(t)-o_m(t-1)
\end{equation}
The loss of first difference voltages at time $t$ is:
\begin{equation}\label{eq:loss_3}
e_{\boldsymbol{w}}(t)=\frac{1}{M} \sum_{m=1}^M \big( \tilde{v}_m(t)- \tilde{o}_m(t) \big)^2 
\end{equation}
Here, $M$ is the number of meters, $\tilde{v}_m(t)=v_m(t)-v_m(t-1)$ is the first difference of actual voltage magnitude measured by meter $m$. In the graphical learning model, we need to calculate the loss function over both the whole data set (i.e., all first difference instances) and mini-batch data (i.e., a smaller set of first difference instances). Thus, we define the gross loss function over a batch of data with time index set $\mathfrak{T}$ as:
\begin{equation}\label{eq:loss_4}
\begin{aligned}
e_{\boldsymbol{w}}(\mathfrak{T}) \triangleq \frac{1}{|\mathfrak{T}|} \sum_{t\in \mathfrak{T}} e_{\boldsymbol{w}}(t)
\end{aligned}
\end{equation}
Here, $|\mathfrak{T}|$ is the size of $\mathfrak{T}$. Suppose we have measurement data over $t=0,...,T$, and define $\mathfrak{T}_{\textrm{full}}\triangleq \{t|t=1,...,T\}$ as the full batch for first difference time series. Then the gross error of the model over all first difference instances is $e_{\boldsymbol{w}}(\mathfrak{T}_{\textrm{full}})$.

\subsection{Gradient of the Loss Function With Respect to the Line Parameters} \label{sec:construct_grad}
We design a new algorithm to calculate the gradient of the loss function \eqref{eq:loss_4} of first difference voltage time series with respect to the line parameters $\boldsymbol{w}$. The gradient calculation formula in the GNN cannot be directly applied because it is derived for the data of a particular time instance, and not for time series. To derive the gradient of the loss function \eqref{eq:loss_4}, we define an equivalent graphical learning model, with new state and feature vectors as follows:
\begin{equation}\label{eq:grad_diff_1}
\hat{\boldsymbol{x}}_n(t) \triangleq 
\begin{bmatrix}
\boldsymbol{x}_n(t-1)  \\
\boldsymbol{x}_n(t)
\end{bmatrix}
,\ 
\hat{\boldsymbol{l}}_n(t) \triangleq 
\begin{bmatrix}
\boldsymbol{l}_n(t-1)  \\
\boldsymbol{l}_n(t)  
\end{bmatrix}
\end{equation}
The corresponding equivalent transition function is:
\begin{equation}\label{eq:grad_diff_2}
\begin{aligned}
\hat{\boldsymbol{x}}_n(t)
& =\hat{f}_{\boldsymbol{w},n}(\hat{\boldsymbol{x}}(t)_n,\hat{\boldsymbol{l}}(t)_n,\hat{\boldsymbol{x}}(t)_{\textrm{ne}(n)}) \\
& \triangleq 
\begin{bmatrix}
f_{\boldsymbol{w},n}(\boldsymbol{x}_n(t-1),\boldsymbol{l}_n(t-1),\boldsymbol{x}_{\textrm{ne}(n)}(t-1))  \\
f_{\boldsymbol{w},n}(\boldsymbol{x}_n(t),\boldsymbol{l}_n(t),\boldsymbol{x}_{\textrm{ne}(n)}(t))
\end{bmatrix}
\end{aligned}
\end{equation}
The compact form of \eqref{eq:grad_diff_2} is:
\begin{equation}\label{eq:grad_diff_3}
[\!\hat{\boldsymbol{x}}(t)\!] \!=\!
\hat{F}_{\boldsymbol{w}}([\!\hat{\boldsymbol{x}}(t)\!],[\hat{\boldsymbol{l}}(t)])
\! \triangleq \!
\begin{bmatrix}
F_{\boldsymbol{w}}([\boldsymbol{x}(t-1)],[\boldsymbol{l}(t-1)])  \\
F_{\boldsymbol{w}}([\boldsymbol{x}(t)],[\boldsymbol{l}(t)]) 
\end{bmatrix}
\end{equation}
Here,
\begin{equation}\label{eq:grad_diff_3_1}
[\hat{\boldsymbol{x}}(t)] \triangleq 
\begin{bmatrix}
[\boldsymbol{x}(t-1)]  \\
[\boldsymbol{x}(t)]
\end{bmatrix},
[\hat{\boldsymbol{l}}(t)] \triangleq 
\begin{bmatrix}
[\boldsymbol{l}(t-1)]  \\
[\boldsymbol{l}(t)]
\end{bmatrix}
\end{equation}
The output function of first difference voltage time series for meter $m$ is:
\begin{equation}\label{eq:grad_diff_4}
\tilde{o}_m(\!t\!)\!=\!
\hat{g}_m(\hat{\boldsymbol{x}}_{\textrm{no}(m)}(\!t\!))
\!\triangleq\!
g_m(\boldsymbol{x}_{\textrm{no}(m)}(\!t\!))-g_m(\boldsymbol{x}_{\textrm{no}(m)}(\!t\!-\!1\!))
\end{equation}
The compact form of \eqref{eq:grad_diff_4} is:
\begin{equation}\label{eq:grad_diff_5}
[\tilde{o}(t)]=\hat{G}([\hat{\boldsymbol{x}}(t)])
\!\triangleq\! 
G([\boldsymbol{x}(t)])-G([\boldsymbol{x}(t-1)])
\end{equation}

Using the equivalent graphical learning model defined in \eqref{eq:grad_diff_1}-\eqref{eq:grad_diff_5}, we can calculate the gradient of $e_{\boldsymbol{w}}(\mathfrak{T})$ over any batch of data $\mathfrak{T}$ with respect to $\boldsymbol{w}$ using an efficient function BACKWARD shown in Algorithm \ref{algorithm:backward}. The iterative FORWARD function can be represented as a recurrent neural network. Thus, $e_{\boldsymbol{w}}(\mathfrak{T})$'s gradient is difficult to calculate in the conventional way. To evaluate the gradient more efficiently, we design Algorithm \ref{algorithm:backward} following the same backpropagation principle in \cite{scarselli2008graph} based on the Almeida-Pineda algorithm \cite{pineda1987generalization,almeida1990learning}. Algorithm \ref{algorithm:backward}  calculates the gradient by using an intermediate variable $\boldsymbol{z}(t)$ through iterative applications of steps 5--10. The theoretical details of designing such algorithms can be found in \cite{scarselli2008graph,pineda1987generalization,almeida1990learning}. In Algorithm \ref{algorithm:backward}, the lengthy derivations of $\hat{A}(t)$, $\hat{\boldsymbol{b}}(t)$, and $\frac{\partial \hat{F}_{\boldsymbol{w}}([\hat{\boldsymbol{x}}(t)],[\hat{\boldsymbol{l}}(t)])}{\partial \boldsymbol{w}}$ are omitted. Please refer to the detailed derivations in Appendix \ref{section:grad_diff_A}, \ref{section:grad_diff_b}, and \ref{section:grad_diff_F}, respectively. $\epsilon_{\textrm{backward}}$ is a small ratio controlling the convergence threshold and $\tilde{\mathfrak{T}}$ is the backward shift batch index defined as:
\begin{equation}\label{eq:grad_diff_6}
\tilde{\mathfrak{T}} \triangleq \{t-1 | t \in \mathfrak{T} \}
\end{equation}

\begin{algorithm}[tbh]
\caption{BACKWARD($\boldsymbol{w}$, $\mathfrak{T}$)} \label{algorithm:backward}
\begin{algorithmic}[1]
\Require Current line parameter $\boldsymbol{w}$ and the first difference instance batch index $\mathfrak{T}$.
\Ensure Gradient $\frac{\partial e_{\boldsymbol{w}}(\mathfrak{T})}{\partial \boldsymbol{w}}$.
\State $[\boldsymbol{x}(t)]$=FORWARD($\boldsymbol{w}$, $t$), $t \in \mathfrak{T} \cup \tilde{\mathfrak{T}}$.
\State Construct $[\hat{\boldsymbol{x}}(t)]$ as \eqref{eq:grad_diff_3_1}, $t \in \mathfrak{T}$.
\State Calculate $[\tilde{o}(t)] =\hat{G}([\hat{\boldsymbol{x}}(t)])$, $\hat{A}(t) =
\frac{\partial \hat{F}_{\boldsymbol{w}}([\hat{\boldsymbol{x}}(t)],[\hat{\boldsymbol{l}}(t)])}{\partial [\hat{\boldsymbol{x}}(t)]}$, $\hat{\boldsymbol{b}}(t) =
\frac{\partial e_{\boldsymbol{w}}(t)}{\partial [\tilde{o}(t)]}\cdot \frac{\partial \hat{G}([\hat{\boldsymbol{x}}(t)])}{\partial [\hat{\boldsymbol{x}}(t)]}$, for $t\in \mathfrak{T}$.
\For{$t\in \mathfrak{T}$}
\State Initialize $\boldsymbol{z}(t)^0=\mathbb{0}_{1\times 12N}$, $\tau=0$.
\Repeat
\State $\boldsymbol{z}(t)^{\tau+1}=\boldsymbol{z}(t)^{\tau}\cdot \hat{A}(t)+\hat{\boldsymbol{b}}(t)$
\State $\tau=\tau+1$
\Until{$\Vert \boldsymbol{z}(t)^{\tau}-\boldsymbol{z}(t)^{\tau-1} \Vert^2 < \epsilon_{\textrm{backward}} \cdot \Vert \boldsymbol{z}(t)^{\tau-1} \Vert^2 $}
\State $\frac{\partial e_{\boldsymbol{w}}(t)}{\partial \boldsymbol{w}}=\boldsymbol{z}(t)^{\tau} \cdot \frac{\partial \hat{F}_{\boldsymbol{w}}([\hat{\boldsymbol{x}}(t)],[\hat{\boldsymbol{l}}(t)])}{\partial \boldsymbol{w}}$, for $t\in \mathfrak{T}$.
\EndFor
\State $\frac{\partial e_{\boldsymbol{w}}(\mathfrak{T})}{\partial \boldsymbol{w}}=\frac{1}{|\mathfrak{T}|} \sum_{t\in \mathfrak{T}} \frac{\partial e_{\boldsymbol{w}}(t)}{\partial \boldsymbol{w}}$
\State \Return $\frac{\partial e_{\boldsymbol{w}}(\mathfrak{T})}{\partial \boldsymbol{w}}$
\end{algorithmic}
\end{algorithm}

\subsection{Utilization of Prior Distribution of Line Parameters Through MAP and Constraints} \label{sec:use_prior}
Electric utilities often have reasonable estimates of distribution systems' line impedance in GIS, which serve as key statistics for the prior distributions of the line parameters. This subsection describes how to use these information to improve estimates of line parameters using maximum a posteriori probability (MAP) and parameter constraints. 

\subsubsection{Use of Prior Line Parameter Distribution in MAP Estimate} \label{sec:MAP}
The posterior distribution of the line parameters is:
\begin{equation}\label{eq:MAP_1}
P(\boldsymbol{w} \ | \ [\tilde{v}(t)]_{t=1}^T)=\frac{P([\tilde{v}(t)]_{t=1}^T \ | \ \boldsymbol{w}) P(\boldsymbol{w})}{P([\tilde{v}(t)]_{t=1}^T)}    
\end{equation}
Here $[\tilde{v}(t)]$ represents a stack of $\tilde{v}_m(t)$, $(m=1,...,M)$, and $[\tilde{v}(t)]_{t=1}^T$ represents $[\tilde{v}(t)]$ of $t=1,...,M$, i.e., the observed first difference voltage time series over the entire time period. Maximizing \eqref{eq:MAP_1} is equivalent to the minimization in \eqref{eq:MAP_2}:
\begin{equation}\label{eq:MAP_2}
\min_{\boldsymbol{w}}  -\log P([\tilde{v}(t)]_{t=1}^T \ | \ \boldsymbol{w})-\log P(\boldsymbol{w})
\end{equation}
We assume $\tilde{v}_m(t) \! \sim \! N(\tilde{o}_m(t),\sigma^2_{v_{m}})$ and are independent across smart meters $m\!=\! 1,...,M$ and time steps $t \! = \! 1,...,T$. We also assume a Gaussian prior of the line parameters $w_i \! \sim \!  N(\mu_i,\sigma^2_{w_i})$, $i\!=\! 1,...,|\boldsymbol{w}|$. $\tilde{o}_m(t)$ is the output of the graphical learning model with parameter $\boldsymbol{w}$, i.e., the theoretical $\tilde{v}_m(t)$ with parameter $\boldsymbol{w}$. For simplification, we further assume $\sigma_{v_{m}} \! \approx \! \sigma_v$, $\forall m$, so that \eqref{eq:MAP_2} can be approximated by:
\begin{equation}\label{eq:MAP_3}
\min_{\boldsymbol{w}} \sum_{t=1}^T \sum_{m=1}^M \frac{(\tilde{v}_m(t)-\tilde{o}_m(t))^2}{\sigma^2_v}
+
\sum_{i=1}^{|\boldsymbol{w}|} \frac{(w_i-\mu_i)^2}{\sigma^2_{w_i}}
\end{equation}
By scaling \eqref{eq:MAP_3}, we have:
\begin{equation}\label{eq:MAP_4}
\begin{aligned}
& \min_{\boldsymbol{w}} \frac{1}{TM}\sum_{t=1}^T \sum_{m=1}^M (\tilde{v}_m(t)-\tilde{o}_m(t))^2
\! + \!
\frac{\sigma^2_v}{TM} \sum_{i=1}^{|\boldsymbol{w}|} \frac{(w_i-\mu_i)^2}{\sigma^2_{w_i}}\\
= & \min_{\boldsymbol{w}} e_{\boldsymbol{w}}(\mathfrak{T}_{\textrm{full}}) + R(\boldsymbol{w})
\end{aligned}
\end{equation}
where $R(\boldsymbol{w}) \triangleq \frac{\sigma^2_v}{TM} \sum_{i=1}^{|\boldsymbol{w}|} \frac{(w_i-\mu_i)^2}{\sigma^2_{w_i}}$. The prior distribution of line parameters specifies, $\mu_i$ and $\sigma^2_{w_i}$. The only unknown term in $R(\boldsymbol{w})$ is $\sigma^2_v$, which needs to be estimated. With the Gaussian assumption $\tilde{v}_m(t)\sim N(\tilde{o}_m(t),\sigma^2_v)$, $\sigma^2_v$ can be estimated from data samples by:
\begin{equation}\label{eq:MAP_5}
\sigma^2_v \! \approx \! \frac{1}{M(T\!-\!1)} \sum_{t=1}^T \sum_{m=1}^M (\tilde{v}_m(t) \! - \! \tilde{o}_m(t))^2 \! = \! \frac{T}{T\!-\!1} e_{\boldsymbol{w}}(\mathfrak{T}_{\textrm{full}})
\end{equation}
The approximation in \eqref{eq:MAP_5} holds when $\boldsymbol{w}$ is close to the true parameter value. The MAP estimation of line parameters consists of two steps. First, we estimate $\boldsymbol{w}$ by minimizing $e_{\boldsymbol{w}}(\mathfrak{T}_{\textrm{full}})$ without prior knowledge and calculate $\sigma^2_v$ with \eqref{eq:MAP_5}. Second, we obtain the MAP estimate with \eqref{eq:MAP_4}.

Since we work with both the entire dataset and mini-batches, we define the loss function over a data batch $\mathfrak{T}$ as:
\begin{equation}\label{eq:MAP_6}
J_{\boldsymbol{w}}(\mathfrak{T})=e_{\boldsymbol{w}}(\mathfrak{T}) + \gamma R(\boldsymbol{w})
\end{equation}
where $\gamma$ is the regularization factor that controls the weight of prior. \eqref{eq:MAP_1}-\eqref{eq:MAP_4} corresponds to MAP with $\gamma=1$. Note that $R(\boldsymbol{w})$ does not depend on $|\mathfrak{T}|$, because $R(\boldsymbol{w})$ is defined on the full batch size $T=|\mathfrak{T}_\textrm{full}|$. This definition ensures that when $\mathfrak{T}_{\textrm{full}}$ is split into mini-batches, the average $J_{\boldsymbol{w}}(\mathfrak{T})$ over all mini-batches equals $J_{\boldsymbol{w}}(\mathfrak{T}_{\textrm{full}})$. 

The gradient of $R(\boldsymbol{w})$ can be calculated as follows:
\begin{equation}\label{eq:MAP_8}
\frac{\partial R(\boldsymbol{w})}{\partial w_i}=\frac{2 \sigma^2_v (w_i-\mu_i)}{TM \sigma^2_{w_i}}, \quad i=1,...,|\boldsymbol{w}|
\end{equation}


\subsubsection{Constraints on Line Parameter Estimates}
We can also add constraints to the line parameter estimates if we know their upper and lower limits. Assume that we know $w_{\textrm{min},i} \leq w_i \leq w_{\textrm{max},i}$, $i,=1,...,|\boldsymbol{w}|$. Then we can apply projected gradient descent to ensure that the learned parameters from the SGD-based estimation procedure stays within the allowable range. Here we denote the projection as $\boldsymbol{w}_{\textrm{proj}}=\textrm{CONS}(\boldsymbol{w}, \boldsymbol{w}_{\textrm{min}}, \boldsymbol{w}_{\textrm{max}})$, in which $w_{\textrm{proj},i}=\min(w_{\textrm{max},i}, \max (w_i, w_{\textrm{min},i}))$ for $i,=1,...,|\boldsymbol{w}|$.


\subsection{SGD-Based Line Parameter Estimation Algorithm} \label{sec:learning_algorithm}
Our proposed SGD-based line parameters estimation method is summarized in Algorithm \ref{algorithm:SGD}. In step 1, the parameter set $\boldsymbol{w}_{\textrm{iter}}$ is initialized with its original value in the GIS. The initial values for the parameters are assumed to be not far from the correct ones. In steps 2 to 20, we iteratively update $\boldsymbol{w_{\textrm{iter}}}$ by descending $J_{\boldsymbol{w}_{\textrm{iter}}}(\mathfrak{T_{\textrm{batch}}})$'s gradient over a small group of samples (i.e., a mini-batch) of size $n_{\textrm{batch}}$. We use patience $n_{\textrm{patience}}$ to decide when to stop the iterative update process. That is to say, the algorithm will be stopped if $J_{\textrm{best}}$ is not improved in $n_{\textrm{patience}}$ epochs (an epoch goes through all $T$ samples in mini-batches). Steps 5 to 15 show the procedure of updating $\boldsymbol{w}_{\textrm{iter}}$ over each mini-batch, in which we use the backtracking line search of parameters $s_{\textrm{initial}}$, $\alpha$, and $\beta$ to determine the step size in each move. In step 21, the parameters $\boldsymbol{w}_{\textrm{best}}$, which has the lowest loss value $J_{\boldsymbol{w}_{\textrm{best}}}(\mathfrak{T_{\textrm{full}}})$ is selected as the output. The use of prior distribution of the distribution line parameters is controlled by $\mu_i$, $\sigma^2_{w_i}$, $i=1,...,|\boldsymbol{w}|$, $\gamma$, $\boldsymbol{w}_{\textrm{min}}$, and $\boldsymbol{w}_{\textrm{max}}$. 

\begin{algorithm}[tbh]
\caption{SGD-Based Line Parameter Estimation} \label{algorithm:SGD}
\begin{algorithmic}[1]
\Require First difference of smart meter voltage magnitude $ [\tilde{v}(t)]$ and three-phase nodal power injection $[\tilde{\boldsymbol{l}}(t)]$, $t\in \mathfrak{T}_{\textrm{full}}$; prior distribution information $\mu_i$, $\sigma^2_{w_i}$ of line parameters, $i=1,...,|\boldsymbol{w}|$, regularization factor $\gamma$, parameter constraints $\boldsymbol{w}_{\textrm{min}}$, $\boldsymbol{w}_{\textrm{max}}$; hyperparameters $n_{\textrm{batch}}$, $n_{\textrm{patience}}$, $s_{\textrm{initial}}$, $\alpha$, $\beta$ and $\epsilon_{stop}$; an initial estimate $\boldsymbol{w}_{\textrm{initial}}$ of $\boldsymbol{w}$ for the $12 \mathfrak{L}$ line parameters.
\Ensure Updated estimate of $\boldsymbol{w}$.
\State Initialize $\boldsymbol{w}_{\textrm{iter}} \! = \! \boldsymbol{w}_{\textrm{best}} \! = \! \boldsymbol{w}_{\textrm{initial}}$ and $J_{\textrm{best}} \!=\! J_{\boldsymbol{w}_{\textrm{best}}}(\mathfrak{T_{\textrm{full}}})$ as \eqref{eq:MAP_6}. $n_{\textrm{epoch}}=0$. $J_{\textrm{history}}(n_{\textrm{epoch}})=J_{\textrm{best}}$.
\Repeat
    \State $n_{\textrm{epoch}}=n_{\textrm{epoch}}+1$
    \State Randomly split $\mathfrak{T}_{\textrm{full}}$ into mini-batches of size $n_{\textrm{batch}}$.
    \For{each mini-batch $\mathfrak{T}_{\textrm{batch}}$}
        \State Calculate $\frac{\partial R(\boldsymbol{w}_{\textrm{iter}})}{\partial \boldsymbol{w}_{\textrm{iter}}}$ as \eqref{eq:MAP_8}.
        \State $\nabla J_{\boldsymbol{w}_{\textrm{iter}}} \!=\! $ BACKWARD($\boldsymbol{w}_{\textrm{iter}}$, $\mathfrak{T}_{\textrm{batch}}$)$+\! \gamma \frac{\partial R(\boldsymbol{w}_{\textrm{iter}})}{\partial \boldsymbol{w}_{\textrm{iter}}}$
        \State Set $s=s_{\textrm{initial}}$ and $\Delta \boldsymbol{w} = -\nabla J_{\boldsymbol{w}_{\textrm{iter}}}$.
        \State $\boldsymbol{w}_{\textrm{temp}}\!=$CONS($\boldsymbol{w}_{\textrm{iter}}+s\Delta \boldsymbol{w}$, $\boldsymbol{w}_{\textrm{min}}$, $\boldsymbol{w}_{\textrm{max}}$)
        \While{$J_{\boldsymbol{w_{\textrm{temp}}}}(\! \mathfrak{T_{\textrm{batch}}} \!) \! > \! J_{\boldsymbol{w}_{\textrm{iter}}}(\! \mathfrak{T_{\textrm{batch}}} \!) \! + \! \alpha s \nabla \! J_{\boldsymbol{w}_{\textrm{iter}}}^T \Delta \boldsymbol{w}$}
            \State $s=\beta s$
            \State $\boldsymbol{w}_{\textrm{temp}}\!= \! \textrm{CONS}(\boldsymbol{w}_{\textrm{iter}}\!+\!s\Delta \! \boldsymbol{w},\boldsymbol{w}_{\textrm{min}},\boldsymbol{w}_{\textrm{max}})$
        \EndWhile
        \State $\boldsymbol{w}_{\textrm{iter}}=\boldsymbol{w}_{\textrm{temp}}$
    \EndFor
    \If{$J_{\boldsymbol{w}_{\textrm{iter}}}(\mathfrak{T_{\textrm{full}}}) < J_{\textrm{best}}$}
        \State $J_{\textrm{best}}=J_{\boldsymbol{w}_{\textrm{iter}}}(\mathfrak{T_{\textrm{full}}}), \boldsymbol{w}_{\textrm{best}}=\boldsymbol{w}_{\textrm{iter}}$.
    \EndIf
    \State $J_{\textrm{history}}(n_{\textrm{epoch}})=J_{\textrm{best}}$
\Until{$1-\frac{J_{\textrm{history}}(n_{\textrm{epoch}})}{J_{\textrm{history}}(n_{\textrm{epoch}}-n_{\textrm{patience}})} < \epsilon_{stop}$}
\State \Return $\boldsymbol{w}_{\textrm{best}}$.
\end{algorithmic}
\end{algorithm}

\subsection{Distributed Parameter Estimation With Network Partition} \label{sec:partition}
For large-scale networks, the FORWARD function takes a larger number iterations to converge and is thus more time consuming. To solve this problem, we propose a network partitioning method to enable parallel computing over smaller sub-networks. The proposed approach works as follows. First, we identify a few edges of the network, which partition the network into sub-networks with similar sizes Second, for each selected edge, one end of it is used as a quasi-source. The quasi-source's three-phase power injection, voltage magnitude of each phase, and the voltage angle difference between phases are measured. Now, each sub-network contains at least one quasi-source node or substation. Third, each sub-network is treated as an independent network and one quasi-source node or substation is selected as the source node; the other quasi-source nodes or substations in this sub-network are treated as ordinary nodes with three additional single-phase pseudo-loads in phase $A$, $B$, and $C$ respectively, whose voltage and power injections are measured. Fourth, we execute Algorithm \ref{algorithm:SGD} for all sub-network in parallel.

\begin{figure}[tbh]
    \centering
    \includegraphics*[width=0.45\textwidth]{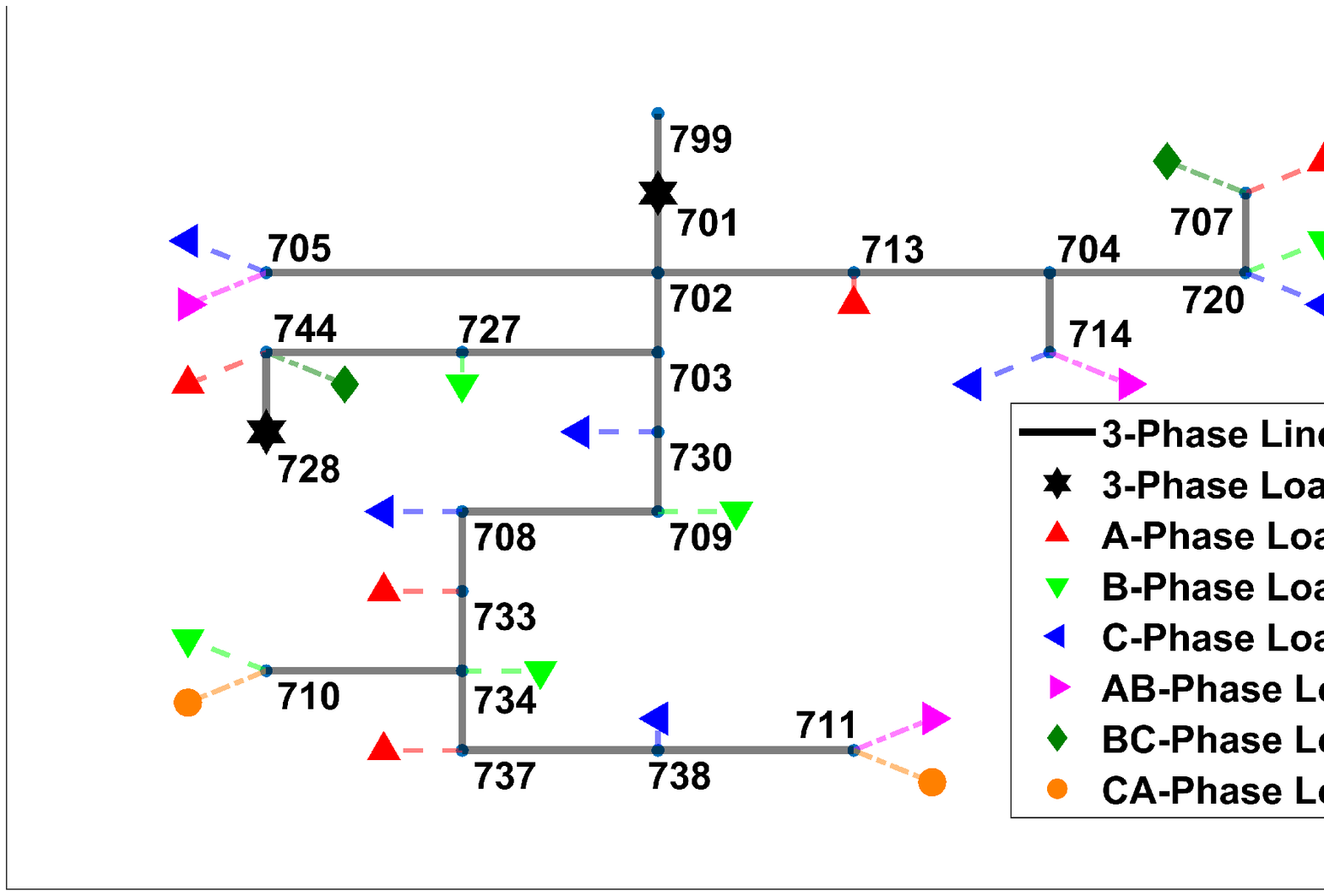}
    \caption{Schematic of the modified IEEE 37-bus test feeder.}
    \label{fig:feeder_37bus_map}
\end{figure}

We can take the IEEE 37-bus test feeder shown in Fig. \ref{fig:feeder_37bus_map} as an example of the network partition method. The feeder is partitioned into three sub-networks with similar size by edge 702-703 and 708-733. Node 702 and 708 are used as quasi-sources. Sub-network 1's source node is 799, and node 702 has 3 additional pseudo loads. Sub-network 2's source node is 702, and node 708 has 3 additional pseudo loads. Sub-network 3's source node is 708. Since sub-network 3 has no other quasi-source nodes, it does not contain any pseudo loads. 

\section{Numerical Study} \label{sec:numerical_study} 
\subsection{Setup for Numerical Tests}
We evaluate the performance of our proposed graphical learning-based parameter estimation algorithm and a few state-of-the-art algorithms on the modified IEEE 13-bus and 37-bus test feeders. We modify these two test feeders by introducing loads with all 7 types of phase connections, $AN$, $BN$, $CN$, $AB$, $BC$, $CA$, and $ABC$. The basic information of the two modified IEEE test feeders are shown in Table \ref{tab:feeder_condition}. The modified 37-bus test feeder is shown in Fig. \ref{fig:feeder_37bus_map} and the modified 13-bus feeder is described in \cite{wang2020parameter}. 
\begin{table}[htb]
\centering
\caption{The Basic Information of the IEEE Test Feeders}
\setlength\tabcolsep{5pt} 
\label{tab:feeder_condition}
\begin{tabular}{c  c  c  c  c}
  \hline
  \hline
  Feeder & \begin{tabular}{@{}c@{}} No. of \\ Loads \end{tabular} & \begin{tabular}{@{}c@{}} No. of \\ Edges \end{tabular} & \begin{tabular}{@{}c@{}} Peak \\ Loads \end{tabular} & \begin{tabular}{@{}c@{}} Level of \\ Unbalance \end{tabular} \\ \hline
  13-bus & 10 & 6 & 3 MW  & 0.0376 \\ 
  37-bus & 25 & 21 & 2.4 MW & 0.0270 \\ \hline \hline
\end{tabular}
\end{table}

The hourly real power consumptions on the test feeders are calculated based on the real power consumption time series from the smart meters of a real-world distribution feeder in North America. The length of the real power consumption time series is 2160, which corresponds to 90 days of measurements. The reactive power time series are calculated by assuming a lagging power factor, which follows a uniform distribution $\mathcal{U}(0.9,1)$. The peak loads of the 13-bus and 37-bus test feeders are 3MW and 2.4MW respectively. The nodal voltages are calculated by power flow analysis using OpenDSS. To simulate the smart meter measurement noise, we use a zero-mean Gaussian distribution with three standard deviation matching 0.1\% to 0.2\% of the nominal values. The 0.1 and 0.2 accuracy class smart meters established in ANSI C12.20-2015 represent the typical noise levels in real-world advanced metering infrastructure. We assume that the initial estimates for the distribution line parameters, $\boldsymbol{w}_{\textrm{initial}}$, are randomly sampled from a uniform distribution within $\pm 50\%$ of the correct values.

When generating simulated time series data, the power consumptions are allocated relatively evenly to each phase so that the test feeders are close to balance. Following \cite{wang2017advanced}, the level of unbalance of a feeder at time interval $t$ can be measured as
\begin{equation}\label{eqCaseStudy-1}
u(t)=\frac{|I_A(t)\!-\!I_m(t)|+|I_B(t)\!-\!I_m(t)|+|I_C(t)\!-\!I_m(t)|}{3 I_m(t)}
\end{equation}
where $I_m(t) = \frac{1}{3} (I_A(t) + I_B(t) + I_C(t))$ is the mean of the distribution substation line current magnitudes of the three phases at time interval $t$. We use the 90-day average of $u(t)$ to measure the level of unbalance of the test feeders, which are shown in Table \ref{tab:feeder_condition}.

The hyperparameters for SGD of the proposed graphical learning model is set up as follows. $n_{\textrm{batch}}\!=\!10$, $n_{\textrm{patience}}\!=\!10$, $s_{\textrm{initial}}\!=\!1000$, $\alpha\!=\!0.3$, $\beta\!=\!0.5$, and $\epsilon_{stop}=0.01$. The $\epsilon_{\textrm{forward}}$ in the FORWARD function and $\epsilon_{\textrm{backward}}$ in the BACKWARD function are set to be $1e-20$. These values are set empirically so that the algorithm updates $J_{\boldsymbol{w}_{\textrm{iter}}}(\mathfrak{T})$ adequately and stops when it saturates. 

The setup corresponding to the prior distribution component of the proposed algorithm is set up as follows. $\boldsymbol{w}_{\textrm{initial}}$ is assumed to be within $\pm 50\%$ of the correct values. Thus, the lower and upper bounds of the parameter $w_i$ are selected to be $\frac{w_{\textrm{initial},i}}{1+50\%}\!=\!\frac{2}{3}w_{\textrm{initial},i}$ and $\frac{w_{\textrm{initial},i}}{1-50\%}\!=\!2w_{\textrm{initial},i}$, where $w_{\textrm{initial},i}$ is the $i$th element in $\boldsymbol{w}_{
\textrm{initial}}$. For the MAP estimation of each parameter $w_i$, we set $\mu_i\!=\!w_{\textrm{initial},i}$ and $\sigma_{w_i}\!=\!w_{\textrm{initial},i}\!\times \! 50 \! \% \times \! \frac{1}{3}$, which represents a Gaussian distribution centered at $w_{\textrm{initial},i}$ and its three standard deviation matching $\pm 50\%$ of $w_{\textrm{initial},i}$. Though this Gaussian assumption is different from the actual uniform distribution of $\boldsymbol{w}_{\textrm{initial}}$, simulation results show the MAP is still effective.

The proposed graphical learning model uses SGD to update line parameter estimates. To reliably evaluate the performance of the proposed model, we execute the algorithm multiple times with different random seeds and calculate the average performance. The numerical tests are implemented using MATLAB on a DELL workstation with two 3.0 GHz Intel Xeon 8-core CPUs and 192 GB RAM.

\subsection{Performance Measurement}
We use the mean absolute deviation ratio (MADR) to measure the estimation error of distribution line parameters. The MADR between the estimated $\boldsymbol{w}$ and the correct value $\boldsymbol{w}^{\dagger}$ is defined as:
\begin{equation} \label{eq:measure-1}
    \text{MADR} \triangleq \sum_{i=1}^{12\mathfrak{L}} | w_i - w^{\dagger}_i| \div \sum_{i=1}^{12\mathfrak{L}} | w^{\dagger}_i| \times 100\%
\end{equation}
The performance of a distribution line parameter estimation algorithm is evaluated by the percentage of MADR improvement, which is defined as: 
\begin{equation} \label{eq:measure-2}
    \textrm{MADR improvement} \triangleq \frac{\textrm{MADR}_{initial}\!-\!\textrm{MADR}_{final}}{\textrm{MADR}_{initial}} \times 100\%
\end{equation}
where $\text{MADR}_{initial}$ and $\text{MADR}_{final}$ represent the MADR of the initial and the final line parameter estimates. The maximum possible MADR improvement is 100\%, which corresponds to a perfect estimation (i.e., $\text{MADR}_{final}\!=\!0\%$).

\subsection{Performance Comparison of the Proposed Graphical Learning Method and State-of-the-Art Algorithms}
The performance of our proposed graphical learning algorithm (GL) with MAP and parameter constraints (abbreviated as CON) is compared with the state-of-the-art algorithm, linearized power flow model based maximum likelihood estimation (LMLE) \cite{wang2020parameter}. In addition, we perform an ablation study to evaluate the relative importance of the MAP and parameter constraints modules in our proposed graphical learning model. These methods are tested with three smart meter accuracy class: noiseless ($0\%$), $0.1\%$, and $0.2\%$. Due to the randomness of the SGD component of the proposed and comparison algorithms, the combination of each algorithm and smart meter class are tested 20 times with different random seeds. The average MADR improvement of the proposed and comparison algorithms are reported in Table \ref{tab:balance}.

\begin{table}[htb]
\centering
\caption{Average MADR Improvement of Parameter Estimation Methods}
\label{tab:balance}
\begin{tabular}{ c c c c c c c}
  \hline
  \hline
  Feeder & \begin{tabular}{@{}c@{}} Meter \\ Class \end{tabular} & LMLE & GL & \begin{tabular}{@{}c@{}} GL+ \\ CON \end{tabular} & \begin{tabular}{@{}c@{}} GL+ \\ MAP \end{tabular} & \begin{tabular}{@{}c@{}} GL+ \\ CON\&MAP \end{tabular} \\ \hline
  \multirow{3}{*}{13-bus} & 0\% & 59.9\% & 69.2\% & 74.7\% & 69.5\% & 75.0\% \\
  & 0.1\% & 59.3\% & 68.0\% & 70.0\% & 70.3\% & 73.4\%  \\
  & 0.2\% & 56.4\% & 64.7\% & 66.4\% & 68.0\% & 70.2\%  \\
  \hline
  \multirow{3}{*}{37-bus} & 0\% & 35.8\% & 40.5\% & 41.7\% & 40.5\% & 41.7\%  \\ 
  & 0.1\% & 17.0\% & 22.0\% & 25.2\% & 25.4\% & 25.6\%  \\
  & 0.2\% & -10.9\% & 10.7\% & 18.7\% & 20.3\% & 20.9\%  \\
  \hline \hline
\end{tabular}
\end{table}

\begin{figure}[tbh]
    \centering
    \includegraphics*[width=0.45\textwidth]{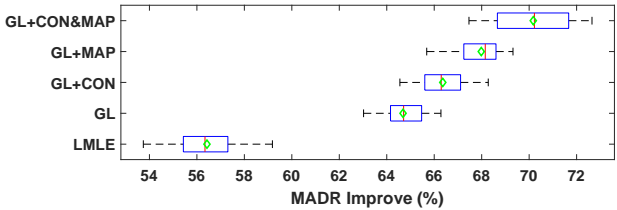}
    \caption{Box plot of 20 random tests for each different algorithms in the 13-bus test feeder, 0.2\% noise level.}
    \label{fig:box_plot}
\end{figure}

From Table \ref{tab:balance}, we can see that the MADR improvement of the GL algorithm is significantly higher than that of LMLE in both test feeders under all meter classes. The increase in MADR improvement ranges from 8.3\% to 9.3\% in the 13-bus feeder and 4.7\% to 21.6\% in the 37-bus feeder. The estimation accuracy of both GL and LMLE increases as the meter noise level decreases. In the 37-bus feeder under 0.2\% meter class, the LMLE has negative MADR improvement, which means the LMLE fails to obtain a more accurate parameter estimation from the initial parameters. On the other hand, the GL algorithm still obtains a more accurate parameter estimation under the same condition. These results show that by preserving the nonlinearity of three-phase power flows, the GL algorithm is significantly more accurate than the LMLE.

In addition to the advantage of GL algorithm, Table \ref{tab:balance} shows the benefit of CON and MAP. Compared with GL algorithm, using only CON has a higher MADR improvement by 1.7\% to 5.5\% in the 13-bus feeder, and 1.2\% to 8\% in the 37-bus feeder. Compared with GL algorithm, using only MAP has a higher MADR improvement by 0.3\% to 3.3\% in the 13-bus feeder, and 0\% to 9.6\% in the 37-bus feeder. The GL algorithm using both CON and MAP has the highest MADR improvement, which is higher than LMLE by 13.8\% to 15.1\% in the 13-bus feeder, and 5.9\% to 31.8\% in the 37-bus feeder. The box plot of Fig. \ref{fig:box_plot} compares the accuracy of different algorithms in the 13-bus test feeder, 0.2\% meter class. These results show that both MAP and CON are effective in utilizing the prior distribution of line parameters to further improve the parameter estimation accuracy.


The estimation accuracy of all algorithms increases as the meter noise level decreases, with one exception. In Table \ref{tab:balance}, we note that under meter class 0\%, the MAP's improvement over the GL algorithm is not as significant as 1\% and 2\% meter classes. This is because under the noiseless 0\% meter class, the $\sigma_v^2$ for MAP is much smaller than 1\% and 2\% meter classes. The smaller $\sigma_v^2$ put less weight on $R(\boldsymbol{w})$ in \eqref{eq:MAP_4} and thus the MAP is less effective under the 0\% meter class.


In Table \ref{tab:balance}, we also note that the overall accuracy of 13-bus feeder is higher than 37-bus. This is because the 37-bus feeder has lower meter number to line number ratio and longer average node-to-node distances (in terms of number of line segments). The material of line segments, configurations, and load profiles are also different between the two feeders.



\subsection{Performance on Unbalanced Distribution Feeders}
We test our proposed method with higher unbalance levels by adjusting the load levels in each phase of the test feeders. The result shows that our proposed method is very accurate even if the feeder is severely unbalanced. Table \ref{tab:unbalance} shows the average MADR improvement of different parameter estimation methods when the feeder's unbalance level is 0.1, which is deemed as severely unbalanced. From  Table \ref{tab:unbalance}, We can draw similar conclusions as in  Table \ref{tab:balance}. The GL algorithm and its combination with CON and MAP significantly outperform LMLE. Compared with the LMLE, the GL algorithm has a higher MADR improvement by 8.4\% to 8.7\% in the 13-bus feeder, and 4.5\% to 19.7\% in the 37-bus feeder. The GL+CON\&MAP has the most accurate estimation result. Its MADR improvement is higher than LMLE by 14.7\% to 16.1\% in the 13-bus feeder, and 6.6\% to 29.5\% in the 37-bus feeder.

\begin{table}[htb]
\centering
\caption{Average MADR Improvement of Parameter Estimation Methods in Highly Unbalanced Feeders(Unbalance Level=0.1)}
\label{tab:unbalance}
\begin{tabular}{ c c c c c c c}
  \hline
  \hline
  Feeder & \begin{tabular}{@{}c@{}} Meter \\ Class \end{tabular} & LMLE & GL & \begin{tabular}{@{}c@{}} GL+ \\ CON \end{tabular} & \begin{tabular}{@{}c@{}} GL+ \\ MAP \end{tabular} & \begin{tabular}{@{}c@{}} GL+ \\ CON\&MAP \end{tabular} \\ \hline
  \multirow{3}{*}{13-bus} & 0\% & 58.2\% & 66.6\% & 73.1\% & 67.7\% & 73.8\% \\ 
  & 0.1\% & 57.6\% & 66.3\% & 68.8\% & 69.6\% & 72.3\%  \\
  & 0.2\% & 54.2\% & 62.9\% & 65.2\% & 67.4\% & 70.3\%  \\
  \hline
  \multirow{3}{*}{37-bus} & 0\% & 35.0\% & 40.4\% & 41.6\% & 40.3\% & 41.6\%  \\
  & 0.1\% & 17.5\% & 22.0\% & 25.3\% & 25.5\% & 25.7\%  \\
  & 0.2\% & -8.4\% & 11.3\% & 19.2\% & 20.7\% & 21.1\%  \\
  \hline \hline
\end{tabular}
\end{table}


\section{Conclusion} \label{sec:conclusion}
In this paper, we develop a physics-informed graphical learning algorithm to estimate line parameters of three-phase power distribution networks. Our proposed algorithm is broadly applicable as it uses only readily available smart meter data to estimate the three-phase series resistance and reactance of the primary line segments. We leverage the domain knowledge of power distribution systems by replacing the deep neural network-based transition functions in the graph neural network with three-phase power flow-based physical transition functions. A rigorous derivation of the gradient of the loss function for first difference voltage time series with respect to line parameters is provided. The network parameters are estimated through iterative application of stochastic gradient descent. The prior distribution of the line parameters is also considered to further improve the accuracy of the proposed parameter estimation algorithm. Comprehensive numerical study results on IEEE test feeders show that our proposed algorithm significantly outperforms the state-of-the-art algorithm. The relative advantage of the proposed algorithm becomes more pronounced when smart meter measurement noise level is higher.

\bibliographystyle{IEEEtran}
\bibliography{Ref_Journal_Para}

\clearpage 
\appendices

\section{Derivation of $\hat{A}(t)$}\label{section:grad_diff_A}
The $12N \!\times \! 12N$ matrix $\hat{A}(t)$ is defined as
\begin{equation}\label{eq:grad_diff_A_1}
\begin{aligned}
\hat{A}(t) & \triangleq
\frac{\partial \hat{F}_{\boldsymbol{w}}([\hat{\boldsymbol{x}}(t)],[\hat{\boldsymbol{l}}(t)])}{\partial [\hat{\boldsymbol{x}}(t)]} \\
&=
\begin{bmatrix}
\frac{\partial F_{\boldsymbol{w}}([\boldsymbol{x}(t-1)],[\boldsymbol{l}(t-1)])}{\partial [\boldsymbol{x}(t-1)]} & \mathbb{0}_{6N\times 6N} \\
\mathbb{0}_{6N\times 6N} & \frac{\partial F_{\boldsymbol{w}}([\boldsymbol{x}(t)],[\boldsymbol{l}(t)])}{\partial [\boldsymbol{x}(t)]}
\end{bmatrix}
\end{aligned}
\end{equation}
$\frac{\partial F_{\boldsymbol{w}}([\boldsymbol{x}(t)],[\boldsymbol{l}(t)])}{\partial [\boldsymbol{x}(t)]}$ is derived by calculating each $6\!\times\!6$ local Jacobian matrix defined as
\begin{equation}\label{eq:grad_diff_A_2}
\frac{\partial f_{\boldsymbol{w},n} (t)}{\partial \boldsymbol{x}_k(t)}
\!\triangleq\!
 \frac{\partial f_{\boldsymbol{w},n}(\!\boldsymbol{x}_n(\!t\!),\boldsymbol{l}_n(\!t\!),\boldsymbol{x}_{\textrm{ne}(\!n\!)}(\!t\!)\!)}{\partial \boldsymbol{x}_k(t)}, \quad 1\! \leq \! n, k \! \leq \! N
\end{equation}
The calculation of \eqref{eq:grad_diff_A_2} depends on $n$ and $k$. If $k \! \notin \! \textrm{ne}(n)$ and $\ k \! \neq \! n$, then
\begin{equation}\label{eq:grad_diff_A_3}
\frac{\partial f_{\boldsymbol{w},n} (t)}{\partial \boldsymbol{x}_k(t)}
=\mathbb{0}_{6\times 6}
\end{equation}
If $k \in \textrm{ne}(n)$, then
\begin{equation}\label{eq:grad_diff_A_4}
\frac{\partial f_{\boldsymbol{w},n} (t)}{\partial \boldsymbol{x}_k(t)}
=\langle Z_{nn} \rangle \cdot \langle Y_{nk} \rangle
\end{equation}
which is a function of line impedance parameters. If $k=n$, then
\begin{equation}\label{eq:grad_diff_A_5}
\frac{\partial f_{\boldsymbol{w},n} (t)}{\partial \boldsymbol{x}_k(t)}
=\langle Z_{nn} \rangle \cdot \frac{\partial \begin{bmatrix}
Re(\boldsymbol{s}_n^*(t)\oslash \boldsymbol{u}_n^*(t))  \\
Im(\boldsymbol{s}_n^*(t)\oslash \boldsymbol{u}_n^*(t)) 
\end{bmatrix}}
{\partial \begin{bmatrix}
Re(\boldsymbol{u}_n(t))  \\
Im(\boldsymbol{u}_n(t)) 
\end{bmatrix}}
\end{equation}
To calculate \eqref{eq:grad_diff_A_5}, we simplify the notations and define
\begin{equation}\label{eq:grad_diff_A_6}
\begin{aligned}
\mathfrak{I}_n^i(t) & \triangleq \frac{p_n^i(t)-jq_n^i(t)}{\alpha_n^i(t)-j\beta_n^i(t)}, \quad i=a,b,c.\\
\end{aligned}
\end{equation}
By rules of the function derivative, each element in the second term of the RHS of \eqref{eq:grad_diff_A_5} can be calculated as in \eqref{eq:grad_diff_A_7} and \eqref{eq:grad_diff_A_8}:
\begin{equation}\label{eq:grad_diff_A_7}
\begin{aligned}
&\frac{\partial Re(\mathfrak{I}_n^i(t))}{\partial \alpha_n^i(t)}
\!=\!\frac{p_n^i(t)[(\beta_n^i(t))^2\!-\!(\alpha_n^i(t))^2]\!-\!2q_n^i(t)\alpha_n^i(t)\beta_n^i(t)}
{[(\alpha_n^i(t))^2+(\beta_n^i(t))^2]^2}\\
&\frac{\partial Re(\mathfrak{I}_n^i(t))}{\partial \beta_n^i(t)}
\!=\!\frac{q_n^i(t)[(\alpha_n^i(t))^2\!-\!(\beta_n^i(t))^2]\!-\!2p_n^i(t)\alpha_n^i(t)\beta_n^i(t)}
{[(\alpha_n^i(t))^2+(\beta_n^i(t))^2]^2}\\
&\frac{\partial Im(\mathfrak{I}_n^i(t))}{\partial \alpha_n^i(t)}
=\frac{\partial Re(\mathfrak{I}_n^i(t))}{\partial \beta_n^i(t)}\\
&\frac{\partial Im(\mathfrak{I}_n^i(t))}{\partial \beta_n^i(t)}
=-\frac{\partial Re(\mathfrak{I}_n^i(t))}{\partial \alpha_n^i(t)}
\end{aligned}
\end{equation}
For $i\neq j$, we have:
\begin{equation}\label{eq:grad_diff_A_8}
\frac{\partial Re(\mathfrak{I}_n^i(t))}{\partial \alpha_n^j(t)}
\!=\!\frac{\partial Re(\mathfrak{I}_n^i(t))}{\partial \beta_n^j(t)}
\!=\!\frac{\partial Im(\mathfrak{I}_n^i(t))}{\partial \alpha_n^j(t)}
\!=\!\frac{\partial Im(\mathfrak{I}_n^i(t))}{\partial \beta_n^j(t)}
\!=\!0
\end{equation}

Thus, given the features $[\hat{\boldsymbol{l}}(t\!-\!1)]$ and $[\hat{\boldsymbol{l}}(t)]$, the line parameter $\boldsymbol{w}$, and the theoretical states $[\hat{\boldsymbol{x}}(t\!-\!1)]$ and $[\hat{\boldsymbol{x}}(t)]$ on current $\boldsymbol{w}$ estimation, we can calculate $\hat{A}(t)$ following \eqref{eq:grad_diff_A_1}-\eqref{eq:grad_diff_A_8}.

\section{Derivation of $\hat{\boldsymbol{b}}(t)$}\label{section:grad_diff_b}
The $1\!\times \! 12N$ vector $\hat{\boldsymbol{b}}(t)$ is defined by
\begin{equation}\label{eq:grad_diff_b_1}
\begin{aligned}
\hat{\boldsymbol{b}}(t) \triangleq
\frac{\partial e_{\boldsymbol{w}}(t)}{\partial [\tilde{o}(t)]}\cdot \frac{\partial \hat{G}([\hat{\boldsymbol{x}}(t)])}{\partial [\hat{\boldsymbol{x}}(t)]}
\end{aligned}
\end{equation}
In \eqref{eq:grad_diff_b_1}, calculating $\frac{\partial e_{\boldsymbol{w}}(t)}{\partial [\tilde{o}(t)]}$ is equivalent to calculating $\frac{\partial e_{\boldsymbol{w}}(t)}{\partial \tilde{o}_m(t)}$, $m\!=\!1,...,M$. From \eqref{eq:loss_3}, we have:
\begin{equation}\label{eq:grad_diff_b_2}
\frac{\partial e_{\boldsymbol{w}}(t)}{\partial \tilde{o}_m(t)}=
\frac{2}{M}\big( \tilde{o}_m(t)-\tilde{v}_m(t) \big), \ m\!=\!1,...,M
\end{equation}
By the definition of \eqref{eq:grad_diff_5}, the second term of RHS of \eqref{eq:grad_diff_b_1} can be calculated as an $M \! \times \! 12N$ matrix:
\begin{equation}\label{eq:grad_diff_b_3}
\frac{\partial \hat{G}([\hat{\boldsymbol{x}}(t)])}{\partial [\hat{\boldsymbol{x}}(t)]}
=
\begin{bmatrix}
-\frac{\partial G([\boldsymbol{x}(t-1)])}{\partial [\boldsymbol{x}(t-1)]}
& \frac{\partial G([\boldsymbol{x}(t)])}{\partial [\boldsymbol{x}(t)]}
\end{bmatrix}
\end{equation}
$\frac{\partial G([\boldsymbol{x}(t)])}{\partial [\boldsymbol{x}(t)]}$ is derived by calculating each $1\!\times\! 6$ vector $\big(\frac{\partial g_m(\boldsymbol{x}_{\textrm{no}(\!m\!)}(t))}{\partial \boldsymbol{x}_n(t)}\big)^T$, $m\!=\!1,...,M$, $n\!=\!1,...,N$. Depending on $m$ and $n$, $\frac{\partial g_m(\boldsymbol{x}_{\textrm{no}(\!m\!)}(t))}{\partial \boldsymbol{x}_n(t)}$ is calculated in three cases. 

\subsubsection{Case 1}
If $n\!\neq \! \textrm{no}(m)$, then:
\begin{equation}\label{eq:grad_diff_b_4}
\frac{\partial g_m(\boldsymbol{x}_{\textrm{no}(m)}(t))}{\partial \boldsymbol{x}_n(t)}=\mathbb{0}_{6\times 1}
\end{equation}

\subsubsection{Case 2}
If $n\!=\!\textrm{no}(m)$ and meter $m$ measures voltage magnitude of phase $i$ (i.e., meter $m$ is a single-phase meter on phase $i$ or a three-phase meter measuring phase $i$'s voltage), then each element of $\frac{\partial g_m(\boldsymbol{x}_{\textrm{no}(m)}(t))}{\partial \boldsymbol{x}_n(t)}$ can be calculated as follows:
\begin{equation}\label{eq:grad_diff_b_5}
\begin{aligned}
\frac{\partial g_m(\boldsymbol{x}_{\textrm{no}(m)}(t))}{\partial \alpha_n^i(t)}
& = \frac{\alpha_n^i(t)}{\sqrt{(\alpha_n^i(t))^2+(\beta_n^i(t))^2}} \\
\frac{\partial g_m(\boldsymbol{x}_{\textrm{no}(m)}(t))}{\partial \beta_n^i(t)}
& = \frac{\beta_n^i(t)}{\sqrt{(\alpha_n^i(t))^2+(\beta_n^i(t))^2}}\\
\frac{\partial g_m(\boldsymbol{x}_{\textrm{no}(m)}(t))}{\partial \alpha_n^j(t)}
& = \frac{\partial g_m(\boldsymbol{x}_{\textrm{no}(m)}(t))}{\partial \beta_n^j(t)}=0 \ (j\neq i)
\end{aligned}
\end{equation}

\subsubsection{Case 3}
If $n=\textrm{no}(m)$ and meter $m$ is a two-phase meter measuring phase $ij$'s voltage magnitude, then each element of $\frac{\partial g_m(\boldsymbol{x}_{\textrm{no}(m)}(t))}{\partial \boldsymbol{x}_n(t)}$ can be calculated as follows:
\begin{equation}\label{eq:grad_diff_b_6}
\begin{aligned}
\frac{\partial g_m(\boldsymbol{x}_{\textrm{no}(m)}(t))}{\partial \alpha_n^i(t)}
& = \frac{\alpha_n^i(t)-\alpha_n^j(t)}{\sqrt{(\alpha_n^i(t)-\alpha_n^j(t))^2+(\beta_n^i(t)-\beta_n^j(t))^2}} \\
\frac{\partial g_m(\boldsymbol{x}_{\textrm{no}(m)}(t))}{\partial \beta_n^i(t)}
& = \frac{\beta_n^i(t)-\beta_n^j(t)}{\sqrt{(\alpha_n^i(t)-\alpha_n^j(t))^2+(\beta_n^i(t)-\beta_n^j(t))^2}} \\
\frac{\partial g_m(\boldsymbol{x}_{\textrm{no}(m)}(t))}{\partial \alpha_n^j(t)}
& = -\frac{\partial g_m(\boldsymbol{x}_{\textrm{no}(m)}(t))}{\partial \alpha_n^i(t)}\\
\frac{\partial g_m(\boldsymbol{x}_{\textrm{no}(m)}(t))}{\partial \beta_n^j(t)}
& = -\frac{\partial g_m(\boldsymbol{x}_{\textrm{no}(m)}(t))}{\partial \beta_n^i(t)}\\
\frac{\partial g_m(\boldsymbol{x}_{\textrm{no}(m)}(t))}{\partial \alpha_n^k(t)}
& = \frac{\partial g_m(\boldsymbol{x}_{\textrm{no}(m)}(t))}{\partial \beta_n^k(t)}=0, \ (k\neq i,j)
\end{aligned}
\end{equation}
Thus, given the theoretical output time difference $\tilde{o}_m(t)$, the measured output time difference $\tilde{v}_m(t)$, and the theoretical states $[\hat{\boldsymbol{x}}(t-1)]$ and $[\hat{\boldsymbol{x}}(t)]$ on current $\boldsymbol{w}$ estimation, we can calculate $\hat{\boldsymbol{b}}(t)$ following \eqref{eq:grad_diff_b_1}-\eqref{eq:grad_diff_b_6}.

\section{Derivation of $\frac{\partial \hat{F}_{\boldsymbol{w}}([\hat{\boldsymbol{x}}(t)],[\hat{\boldsymbol{l}}(t)])}{\partial \boldsymbol{w}}$}\label{section:grad_diff_F}
From \eqref{eq:grad_diff_3}, we have the $12N \! \times \! 12\mathfrak{L}$ matrix
\begin{equation}\label{eq:grad_diff_F_1}
\frac{\partial \hat{F}_{\boldsymbol{w}}([\hat{\boldsymbol{x}}(t)],[\hat{\boldsymbol{l}}(t)])}{\partial \boldsymbol{w}}
=
\begin{bmatrix}
\frac{\partial F_{\boldsymbol{w}}([\boldsymbol{x}(t-1)],[\boldsymbol{l}(t-1)])}{\partial \boldsymbol{w}}  \\
\frac{\partial F_{\boldsymbol{w}}([\boldsymbol{x}(t)],[\boldsymbol{l}(t)])}{\partial \boldsymbol{w}} 
\end{bmatrix}
\end{equation}
$\frac{\partial F_{\boldsymbol{w}}([\boldsymbol{x}(t)],[\boldsymbol{l}(t)])}{\partial \boldsymbol{w}}$ is derived by calculating $\frac{\partial f_{\boldsymbol{w},n} (t)}{\partial w_m}$ for each $n\!=\!1,...,N$ and $m\!=\!1,....|\boldsymbol{w}|$, in which
\begin{equation}\label{eq:grad_diff_F_2}
f_{\boldsymbol{w},n} (t)
\triangleq
f_{\boldsymbol{w},n}(\boldsymbol{x}_n(t),\boldsymbol{l}_n(t),\boldsymbol{x}_{\textrm{ne}(n)}(t))
\end{equation}
For easier derivation, here we introduce a new set of parameters $\boldsymbol{\xi}$ of size $12\mathfrak{L}$, which is the set of $\boldsymbol{w}$'s corresponding line conductance and susceptance. Then $\frac{\partial f_{\boldsymbol{w},n} (t)}{\partial w_m}$ is derived by
\begin{equation}\label{eq:grad_diff_F_3}
\frac{\partial f_{\boldsymbol{w},n} (t)}{\partial w_m}
=\frac{\partial f_{\boldsymbol{w},n} (t)}{\partial \boldsymbol{\xi}}
\cdot
\frac{\partial \boldsymbol{\xi}}{\partial w_m}
\end{equation}
$\frac{\partial f_{\boldsymbol{w},n} (t)}{\partial \boldsymbol{\xi}}$ is calculated by calculating each $\frac{\partial f_{\boldsymbol{w},n} (t)}{\partial \xi_m}$, $m\!=\! 1,...,|\boldsymbol{\xi}|$. From \eqref{eq:transition-4}, we have
\begin{equation}\label{eq:grad_diff_F_4}
\begin{aligned}
\frac{\partial f_{\boldsymbol{w},n} (t)}{\partial \xi_m} & =
\frac{\partial \langle Z_{nn} \rangle}{\partial \xi_m}
\bigg(\!
\begin{bmatrix}
Re(\boldsymbol{s}_n^*(t)\oslash \boldsymbol{u}_n^*(t))  \\
Im(\boldsymbol{s}_n^*(t)\oslash \boldsymbol{u}_n^*(t)) 
\end{bmatrix}\\
& +\!\sum_{k\in \textrm{ne}(n)} \! \langle Y_{nk} \rangle \!
\begin{bmatrix}
Re(\boldsymbol{u}_k(t))  \\
Im(\boldsymbol{u}_k(t)) 
\end{bmatrix}
\! \bigg)\\
& + \langle Z_{nn} \rangle \sum_{k\in \textrm{ne}(n)} \! \frac{\partial \langle Y_{nk} \rangle}{\partial \xi_m}
\begin{bmatrix}
Re(\boldsymbol{u}_k(t))  \\
Im(\boldsymbol{u}_k(t)) 
\end{bmatrix}
\end{aligned}
\end{equation}
\eqref{eq:grad_diff_F_3} and \eqref{eq:grad_diff_F_4} can be calculated given current parameter estimate $\boldsymbol{w}$, corresponding $\boldsymbol{\xi}$, and the theoretical state $[\hat{\boldsymbol{x}}(t)]$ on current $\boldsymbol{w}$ estimation. The derivation of $\frac{\partial \langle Z_{nn} \rangle}{\partial \xi_m}$ and $\frac{\partial \langle Y_{nk} \rangle}{\partial \xi_m}$ in \eqref{eq:grad_diff_F_4} will be explained in Appendix section \ref{sec:d_Z_d_xi}. The derivation of $\frac{\partial \boldsymbol{\xi}}{\partial w_m}$ in \eqref{eq:grad_diff_F_3} will be explained in Appendix section \ref{sec:dev_dxi_dw}.

\subsection{Derivation of $\frac{\partial \langle Z_{nn} \rangle}{\partial \xi_m}$ and $\frac{\partial \langle Y_{nk} \rangle}{\partial \xi_m}$}\label{sec:d_Z_d_xi}
From \eqref{eq:transition-3}, we have
\begin{equation}\label{eq:grad_diff_F_5}
\frac{\partial \langle Z_{nn} \rangle}{\partial \xi_m}
=
\begin{bmatrix}
\frac{\partial Re(Z_{nn})}{\partial \xi_m} & -\frac{\partial Im(Z_{nn})}{\partial \xi_m} \\
\frac{\partial Im(Z_{nn})}{\partial \xi_m} & \frac{\partial Re(Z_{nn})}{\partial \xi_m}
\end{bmatrix}
\end{equation}
By the definition in \eqref{eq:transition-1} and \eqref{eq:transition-4}, we have
\begin{equation}\label{eq:grad_diff_F_6}
Z_{nn}=Y_{nn}^{-1}=(G_{nn}+jB_{nn})^{-1}
\end{equation}
Here, $G_{nn}\!=\!\sum_{k\in \textrm{ne}(n)} G_{nk}$ and $B_{nn}\!=\!\sum_{k\in \textrm{ne}(n)} B_{nk}$. $G_{nk}$ and $B_{nk}$ are the real and imaginary part of $Y_{nk}$. For a complex square matrix $(A+jB)$, if $A$ and $(A+BA^{-1}B)$ are nonsingular, then by the Woodbury matrix identity, we can prove the following:
\begin{equation}\label{eq:grad_diff_F_7}
(\!A\!+\!jB\!)^{-1}\!=\!(\!A\!+\!BA^{-1}B\!)^{-1}\!-\!j(\!A\!+\!BA^{-1}B\!)^{-1}BA^{-1}
\end{equation}
Under normal conditions, the $G_{nn}$ and $B_{nn}$ satisfy the condition for \eqref{eq:grad_diff_F_7}, which is also verified by numerical tests. Thus, we have
\begin{equation}\label{eq:grad_diff_F_8}
\begin{aligned}
\frac{\partial Re(Z_{nn})}{\partial \xi_m}
& =
\frac{\partial (G_{nn}+B_{nn}G_{nn}^{-1}B_{nn})^{-1}}{\partial \xi_m}\\
\frac{\partial Im(Z_{nn})}{\partial \xi_m}
& =
-\frac{\partial (G_{nn}+B_{nn}G_{nn}^{-1}B_{nn})^{-1}B_{nn}G_{nn}^{-1}}{\partial \xi_m}
\end{aligned}
\end{equation}
The $3\! \times \!3$ matrix $\frac{\partial Re(\!Z_{nn}\!)}{\partial \xi_m}$ is derived by calculating $\frac{\partial Re(Z_{nn}(i,j))}{\partial \xi_m}$ for each $i,j$, in which $Z_{nn}(i,j)$ is the $ij$-th element of $Z_{nn}$. By the chain rule, we have
\begin{equation}\label{eq:grad_diff_F_9}
\begin{aligned}
\frac{\partial Re(Z_{nn}(i,j))}{\partial \xi_m}
\!=\! \Tr \! \bigg(\! \bigg[ \! \frac{\partial Re(Z_{nn}(i,j))}{\partial (Re(Z_{nn}))^{-1}} \! \bigg]^T
\! \times \! 
\frac{\partial (Re(Z_{nn}))^{-1}}{\partial \xi_m}
\!\bigg)
\end{aligned}
\end{equation}
We define $E^{(i,j)}_{m \times n}$ as an $m\times n$ matrix, in which the $ij$-th element is 1 and the rest of elements are all 0. Using the rules of matrix derivatives \cite{zwillinger2018crc}, we have
\begin{equation}\label{eq:grad_diff_F_10}
\begin{aligned}
\frac{\partial Re(Z_{nn}(i,j))}{\partial (Re(Z_{nn}))^{-1}}
=
-Re(Z_{nn})^T E_{3\times 3}^{(i,j)} Re(Z_{nn})^T
\end{aligned}
\end{equation}
The second term of RHS of \eqref{eq:grad_diff_F_9} is calculated following \eqref{eq:grad_diff_F_8}:
\begin{equation}\label{eq:grad_diff_F_11}
\begin{aligned}
& \frac{\partial (Re(Z_{nn}))^{-1}}{\partial \xi_m}=\frac{\partial (G_{nn}+B_{nn}G_{nn}^{-1}B_{nn})}{\partial \xi_m}\\
=& \frac{\partial G_{nn}}{\partial \xi_m}
\!+\!\frac{\partial B_{nn}}{\partial \xi_m} G_{nn}^{-1}B_{nn}
\!+\!B_{nn} \frac{\partial G_{nn}^{-1}}{\partial \xi_m} B_{nn} \\
+ & B_{nn} G_{nn}^{-1} \frac{\partial B_{nn}}{\partial \xi_m}
\end{aligned}
\end{equation}
Here $\frac{\partial G_{nn}}{\partial \xi_m}\!=\!\sum_{k\in \textrm{ne}(n)} \frac{\partial G_{nk}}{\partial \xi_m}$ and $\frac{\partial B_{nn}}{\partial \xi_m}\!=\! \sum_{k\in \textrm{ne}(n)} \frac{\partial B_{nk}}{\partial \xi_m}$. Calculating $\frac{\partial G_{nk}}{\partial \xi_m}$ and $\frac{\partial B_{nk}}{\partial \xi_m}$ is straight forward as in \eqref{eq:grad_diff_F_12} and \eqref{eq:grad_diff_F_13}.
\begin{equation}\label{eq:grad_diff_F_12}
\frac{\partial G_{nk}}{\partial \xi_m}
\!=\!
\begin{cases}\!
\begin{aligned}
 & \mathbb{0}_{3\! \times \! 3} \ \textrm{if $\xi_m$ is not line $nk$'s conductance parameter} \\
 & E^{(i,i)}_{3 \times 3} \ \textrm{if $\xi_m$ is the $ii$-th diagonal element in $G_{nk}$} \\
 & E^{(i,j)}_{3 \! \times \! 3}\!+\!E^{(j,i)}_{3 \! \times \! 3} \ \textrm{if $\xi_m$ is the $ij$-th and $ji$-th} \\
 & \quad\qquad\qquad\qquad \textrm{off-diagonal elements in $G_{nk}$}
\end{aligned}
\end{cases}
\end{equation}
\begin{equation}\label{eq:grad_diff_F_13}
\frac{\partial B_{nk}}{\partial \xi_m}
\!=\!
\begin{cases}
\begin{aligned}
 & \mathbb{0}_{3\! \times \! 3} \ \textrm{if $\xi_m$ is not line $nk$'s susceptance parameter} \\
 & E^{(i,i)}_{3\! \times \!3} \ \textrm{if $\xi_m$ is the $ii$-th diagonal element in $B_{nk}$} \\
 & E^{(i,j)}_{3\! \times \! 3}\!+\!E^{(j,i)}_{3\! \times \!3} \ \textrm{if $\xi_m$ is the $ij$-th and $ji$-th} \\
 & \quad\qquad\qquad\qquad \textrm{off-diagonal elements in $B_{nk}$}
\end{aligned}
\end{cases}
\end{equation}
The $3\times 3$ matrix $\frac{\partial G_{nn}^{-1}}{\partial \xi_m}$ is derived by calculating $\frac{\partial G_{nn}^{-1}(i,j)}{\partial \xi_m}$ for each $i$, $j$, in which $G_{nn}^{-1}(i,j)$ is the $ij$-th element of $G_{nn}^{-1}$. By the chain rule, we have
\begin{equation}\label{eq:grad_diff_F_14}
\begin{aligned}
\frac{\partial G_{nn}^{-1}(i,j)}{\partial \xi_m}
= \Tr \bigg( \bigg[ \frac{\partial G_{nn}^{-1}(i,j)}{\partial G_{nn}} \bigg]^T
\times 
\frac{\partial G_{nn}}{\partial \xi_m}
\bigg)
\end{aligned}
\end{equation}
We have shown how to calculate $\frac{\partial G_{nn}}{\partial \xi_m}$. And similar to \eqref{eq:grad_diff_F_10}, we have
\begin{equation}\label{eq:grad_diff_F_15}
\begin{aligned}
\frac{\partial G_{nn}^{-1}(i,j)}{\partial G_{nn}}
=
- G_{nn}^{-T} E_{3\times 3}^{(i,j)} G_{nn}^{-T}
\end{aligned}
\end{equation}
From \eqref{eq:grad_diff_F_8}, we have
\begin{equation}\label{eq:grad_diff_F_18}
\begin{aligned}
\frac{\partial Im(Z_{nn})}{\partial \xi_m}
& =
-\frac{\partial Re(Z_{nn})}{\partial \xi_m}B_{nn}G_{nn}^{-1}
-Re(Z_{nn})\frac{\partial B_{nn}}{\partial \xi_m}G_{nn}^{-1}\\
& -Re(Z_{nn})B_{nn}\frac{\partial G_{nn}^{-1}}{\partial \xi_m}
\end{aligned}
\end{equation}
Every term in \eqref{eq:grad_diff_F_18} has been solved by \eqref{eq:grad_diff_F_9}-\eqref{eq:grad_diff_F_15}.

The $\frac{\partial \langle Y_{nk} \rangle}{\partial \xi_m}$ in \eqref{eq:grad_diff_F_4} can be calculated as
\begin{equation}\label{eq:grad_diff_F_19}
\frac{\partial \langle \! Y_{nk} \!\rangle}{\partial \xi_m}
\!=\!
\begin{bmatrix}
\frac{\partial Re(\!Y_{nk}\!)}{\partial \xi_m} & -\frac{\partial Im(\!Y_{nk}\!)}{\partial \xi_m}\\
\frac{\partial Im(\!Y_{nk}\!)}{\partial \xi_m} & \frac{\partial Re(\!Y_{nk}\!)}{\partial \xi_m}
\end{bmatrix}
\!=\!
\begin{bmatrix}
\frac{\partial G_{nk}}{\partial \xi_m} & -\frac{\partial B_{nk}}{\partial \xi_m} \\
\frac{\partial B_{nk}}{\partial \xi_m} & \frac{\partial G_{nk}}{\partial \xi_m} 
\end{bmatrix}
\end{equation}
Here, every element in \eqref{eq:grad_diff_F_19} can be calculated by \eqref{eq:grad_diff_F_12} and \eqref{eq:grad_diff_F_13}.

\subsection{Derivation of $\frac{\partial \boldsymbol{\xi}}{\partial w_m}$}\label{sec:dev_dxi_dw}
Since $\boldsymbol{\xi}$ is the set of $12\mathfrak{L}$ lines' conductance and susceptance, we have $\boldsymbol{\xi}\!=\!\{g_l^{ij}, b_l^{ij} \, | \, l\!=\!1,...,12\mathfrak{L}, ij\!=\!aa,ab,ac,bb,bc,cc\}$, in which $g_l^{ij}$ and $b_l^{ij}$ are line $l$'s conductance and susceptance in phase $ij$. Thus, we need to calculate $\frac{\partial g_l^{ij}}{\partial w_m}$ and $\frac{\partial b_l^{ij}}{\partial w_m}$. Let $G_l$ and $B_l$ be the $3\! \times \!3$ conductance and susceptance matrix of line $l$. From \eqref{eq:grad_diff_F_7}, we know
\begin{equation}\label{eq:grad_diff_w_3}
\begin{aligned}
G_l &=(R_l+X_l R_l^{-1} X_l)^{-1}\\
B_l &=-G_l X_l R_l^{-1}
\end{aligned}
\end{equation}
By the chain rule, we have
\begin{equation}\label{eq:grad_diff_w_4}
\begin{aligned}
\frac{\partial g_l^{ij}}{\partial w_m}
& = \Tr \bigg( \bigg[ \frac{\partial g_l^{ij}}{\partial G_l^{-1}} \bigg]^T
\times 
\frac{\partial G_l^{-1}}{\partial w_m}\bigg)\\
\frac{\partial b_l^{ij}}{\partial w_m}
& = \Tr \bigg( \bigg[ \frac{\partial b_l^{ij}}{\partial B_l^{-1}} \bigg]^T
\times 
\frac{\partial B_l^{-1}}{\partial w_m}
\bigg)
\end{aligned}
\end{equation}
Suppose $g_l^{ij}$ and $b_l^{ij}$ are the $hk$-th element of $G_l$ and $B_l$, ($h\leq k$). Similar to \eqref{eq:grad_diff_F_10}, we have
\begin{equation}\label{eq:grad_diff_w_5}
\begin{aligned}
\frac{\partial g_l^{ij}}{\partial G_l^{-1}}
& =-G_l^T E^{(h,k)}_{3 \times 3} G_l^T \\
\frac{\partial b_l^{ij}}{\partial B_l^{-1}}
& =-B_l^T E^{(h,k)}_{3 \times 3} B_l^T
\end{aligned}
\end{equation}
From \eqref{eq:grad_diff_w_3}, we have:
\begin{equation}\label{eq:grad_diff_w_6}
\begin{aligned}
\frac{\partial G_l^{-1}}{\partial w_m}
&=
\frac{\partial R_l}{\partial w_m}
+
\frac{\partial X_l}{\partial w_m} R_l^{-1} X_l \\
& + X_l \frac{\partial R_l^{-1}}{\partial w_m} X_l
+ X_l R_l^{-1} \frac{\partial X_l}{\partial w_m}\\
\frac{\partial B_l^{-1}}{\partial w_m}
&=
-\frac{\partial G_l}{\partial w_m} X_l R_l^{-1}
-G_l \frac{\partial X_l}{\partial w_m} R_l^{-1}\\
& - G_l X_l \frac{\partial R_l^{-1}}{\partial w_m}
\end{aligned}
\end{equation}
Here,
\begin{equation}\label{eq:grad_diff_w_7}
\frac{\partial R_l}{\partial w_m}
\!=\!
\begin{cases}
\begin{aligned}
 & \mathbb{0}_{3 \times 3} \ \textrm{if $w_m$ is not line $l$'s resistance parameter} \\
 & E^{(d,d)}_{3 \times 3} \ \textrm{if $w_m$ is the $dd$-th diagonal element in $R_l$} \\
 & E^{(d,e)}_{3 \times 3}\!+\!E^{(e,d)}_{3 \times 3} \ \textrm{if $w_m$ is the $de$-th and $ed$-th} \\
 & \quad\qquad\qquad\qquad \textrm{off-diagonal elements in $R_l$}
\end{aligned}
\end{cases}
\end{equation}

\begin{equation}\label{eq:grad_diff_w_8}
\frac{\partial X_l}{\partial w_m}
\!=\!
\begin{cases}
\begin{aligned}
 & \mathbb{0}_{3 \times 3} \ \textrm{if $w_m$ is not line $l$'s reactance parameter} \\
 & E^{(d,d)}_{3 \times 3} \ \textrm{if $w_m$ is the $dd$-th diagonal element in $X_l$} \\
 & E^{(d,e)}_{3 \times 3}\!+\!E^{(e,d)}_{3 \times 3} \ \textrm{if $w_m$ is the $de$-th and $ed$-th} \\
 & \quad\qquad\qquad\qquad \textrm{off-diagonal elements in $X_l$}
\end{aligned}
\end{cases}
\end{equation}
$\frac{\partial R_l^{-1}}{\partial w_m}$ is derived by calculating $\frac{\partial R_l^{-1}(d,e)}{\partial w_m}$ for each $d$ and $e$, where $R_l^{-1}(d,e)$ is the $de$-th element of $R_l^{-1}$.
\begin{equation}\label{eq:grad_diff_w_9}
\begin{aligned}
\frac{\partial R_l^{-1}(d,e)}{\partial w_m}
= \Tr \bigg( \bigg[ \frac{\partial R_l^{-1}(d,e)}{\partial R_l} \bigg]^T
\times 
\frac{\partial R_l}{\partial w_m}
\bigg)
\end{aligned}
\end{equation}
And similar to \eqref{eq:grad_diff_F_10}, we have
\begin{equation}\label{eq:grad_diff_w_10}
\begin{aligned}
\frac{\partial R_l^{-1}(d,e)}{\partial R_l}
& =-R_l^{-T} E^{(d,e)}_{3 \times 3} R_l^{-T} 
\end{aligned}
\end{equation}
Thus, following \eqref{eq:grad_diff_w_3}-\eqref{eq:grad_diff_w_10}, we derive $\frac{\partial \boldsymbol{\xi}}{\partial w_m}$. Plugging it into \eqref{eq:grad_diff_F_3}, we calculate $\frac{\partial f_{\boldsymbol{w},n} (t)}{\partial w_m}$ and thus obtain $\frac{\partial \hat{F}_{\boldsymbol{w}}([\hat{\boldsymbol{x}}(t)],[\hat{\boldsymbol{l}}(t)])}{\partial \boldsymbol{w}}$.
\end{document}